\documentclass[letterpaper, 10 pt, journal, twoside]{IEEEtran}

\usepackage{cite}
\usepackage{times}
\usepackage{graphicx}
\usepackage{amsmath}
\usepackage{amssymb}
\usepackage{subcaption}
\usepackage{gensymb}
\usepackage{booktabs}
\usepackage[style=base]{caption}
\usepackage{mathtools}
\usepackage{pifont}
\usepackage{xspace}
\usepackage{xfrac}
\usepackage{microtype}
\usepackage[table]{xcolor}

\definecolor{darkgreen}{RGB}{0, 150, 0}
\definecolor{darkred}{RGB}{200, 0, 0}
\definecolor{darkblue}{RGB}{0, 0, 200}

\ifx00 %ifx01 to turn off and ifx00 to turn on
\newcommand{\sy}[1]{\textcolor{red}{$<$SY: #1$>$}}
\newcommand{\vr}[1]{\textcolor{orange}{$<$VR: #1$>$}}
\newcommand{\ce}[1]{\textcolor{blue}{$<$CE: #1$>$}}
\else
\newcommand{\sy}[1]{}
\newcommand{\vr}[1]{}
\newcommand{\ce}[1]{}
\fi
\usepackage[utf8]{inputenc}
\usepackage{bm}
\newcommand{\matr}[1]{#1}               % pure math version
\renewcommand{\vec}[1]{\mathbf{#1}}     % vector

\newcommand{\spce}[1]{\mathbb{#1}}

\newcommand{\T}{\mathsf{T}}  % Preferred to \top
\newcommand{\atan}{\text{atan}}
\newcommand{\acos}{\text{acos}}
\newcommand{\asin}{\text{asin}}
\newcommand{\onimg}{\tau}  % character for on-image function
\newcommand{\prjt}{\pi}  % character for projection function
\newcommand{\radl}{\varrho}  % character for radial function
\newcommand{\mat}[1]{\ensuremath{\begin{pmatrix}#1\end{pmatrix}}}

\makeatletter
\let\NAT@parse\undefined
\DeclareRobustCommand\onedot{\futurelet\@let@token\@onedot}
\def\@onedot{\ifx\@let@token.\else.\null\fi\xspace}
\def\eg{\emph{e.g}\onedot} 
\def\ie{\emph{i.e}\onedot} 
\def\etal{\emph{et al}\onedot}
\makeatother

\definecolor{lightgray}{RGB}{131, 126 114}
\definecolor{darkgreen}{RGB}{0, 150, 0}
\definecolor{darkred}{RGB}{200, 0, 0}
\definecolor{darkblue}{RGB}{0, 0, 200}
\definecolor{gray}{RGB}{142,142,142}
\newcommand{\ch}{{\color{darkgreen} \ding{51}}}
\newcommand{\xm}{{\color{darkred} \ding{55}}}

\usepackage[colorlinks,pagebackref=false,citecolor=blue,bookmarks=false,hypertexnames=true]{hyperref}
\begin{document}

\title{Surround-view Fisheye Camera Perception for Automated Driving: Overview, Survey \& Challenges}

\author{Varun~Ravi~Kumar,
        Ciar\'{a}n Eising~\IEEEmembership{Member,~IEEE},
        Christian Witt, and
        Senthil~Yogamani
\thanks{V. Ravi Kumar, C. Witt, and S. Yogamani are with Valeo.}
\thanks{C. Eising is with the Dept. of Electronic and Computer Engineering at the University of Limerick.}
\thanks{V. Ravi Kumar and C. Eising are co-first authors.}
\thanks{(Corresponding author: Ciar\'{a}n Eising)  (email: ciaran.eising@ul.ie)}
%\thanks{Manuscript received December 25th, 2022}
}
% -------------------------------------------------
%\markboth{IEEE Transactions on Intelligent Transportation Systems}
%{}
\maketitle
% -------------------------------------------------
% Abstract
% -------------------------------------------------
\begin{abstract}
Surround-view fisheye cameras are commonly used for near-field sensing in automated driving. Four fisheye cameras on four sides of the vehicle are sufficient to cover $360\degree$ around the vehicle capturing the entire near-field region. Some primary use cases are automated parking, traffic jam assist, and urban driving. There are limited datasets and very little work on near-field perception tasks as the focus in automotive perception is on far-field perception. In contrast to far-field, surround-view perception poses additional challenges due to high precision object detection requirements of 10cm and partial visibility of objects. Due to the large radial distortion of fisheye cameras, standard algorithms cannot be extended easily to the surround-view use case. Thus, we are motivated to provide a self-contained reference for automotive fisheye camera perception for researchers and practitioners. Firstly, we provide a unified and taxonomic treatment of commonly used fisheye camera models. Secondly, we discuss various perception tasks and existing literature. Finally, we discuss the challenges and future direction.
\end{abstract}
% -------------------------------------------------
\begin{IEEEkeywords}
Automated Driving, Omnidirectional Camera, Fisheye Camera, Surround View Perception, Bird-eye's View Perception, Multi-Task Learning
\end{IEEEkeywords}
% Intro 
% -------------------------------------------------
\section{Introduction} 
\label{sec:intro}

\begin{figure}[t]
\centering
 \captionsetup{font=footnotesize, belowskip=0pt}
\includegraphics[width=0.85\columnwidth,page=2]{images/Surround_view.pdf}

\vspace{1mm} % don't delete white space above and below, it causes issues

\includegraphics[width=0.85\columnwidth,page=2]{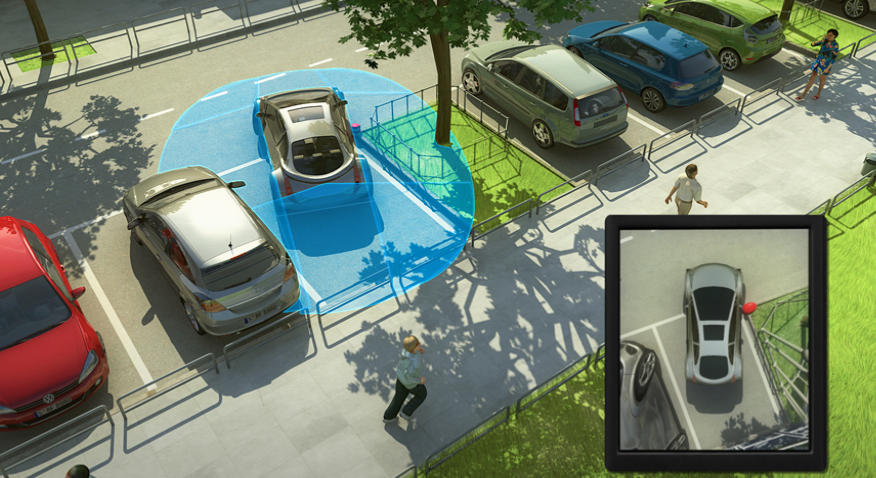}
\caption{\textbf{Illustration of a typical automotive surround-view system} consisting of four fisheye cameras located at the front, rear, and on each wing mirror (top). The bottom figure illustrates the near field covering the entire $360^\circ$ around the vehicle. Surround visualization for the driver by stitching the four cameras is also illustrated within the smaller box.}
\label{fig:surroundview}
\end{figure}

Surround-view systems use four sensors to form a network with overlap regions, sufficient to cover the near-field area around the car. Figure \ref{fig:surroundview} shows the four views of a typical surround-view system, along with a representation of the typical parking use-case. Wide-angle views exceeding $180^\circ$ are used for this near-field sensing. Any perception algorithm must consider the significant fisheye distortion inherent with such camera systems. This is a significant challenge, as most work in computer vision focuses on narrow field-of-view cameras with mild radial distortion. However, as such camera systems are more widely deployed, work has been completed in this area. It is the aim of this paper to give the reader an overview of surround view cameras (\eg, image formation, configuration, and rectification), to survey the existing state of the art, and to provide insights into the current challenges in the area.

\begin{figure}[t]
\centering
 \captionsetup{font=footnotesize, belowskip=0pt}
\includegraphics[width=0.75\linewidth]{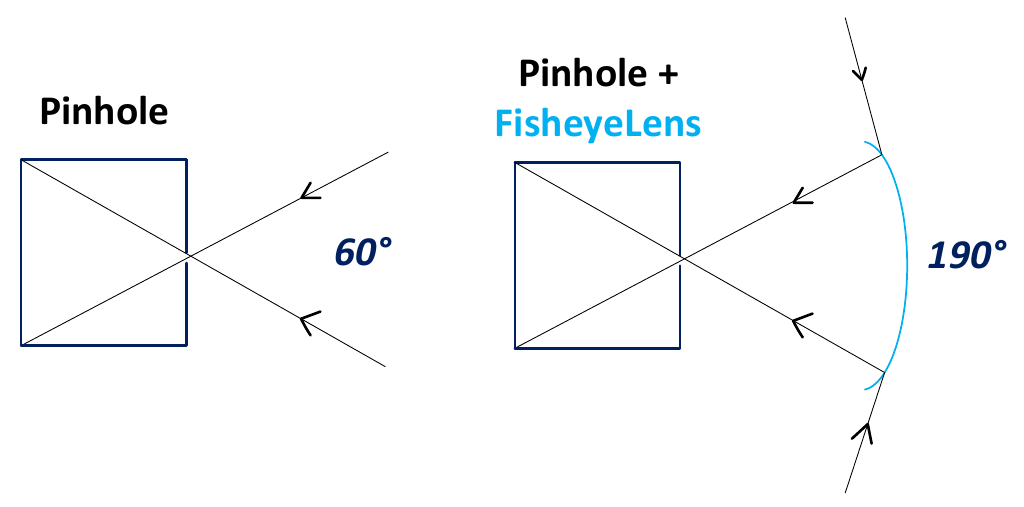}

\vspace{1mm} % don't delete white space above and below, it causes issues

  \includegraphics[width=0.75\linewidth]{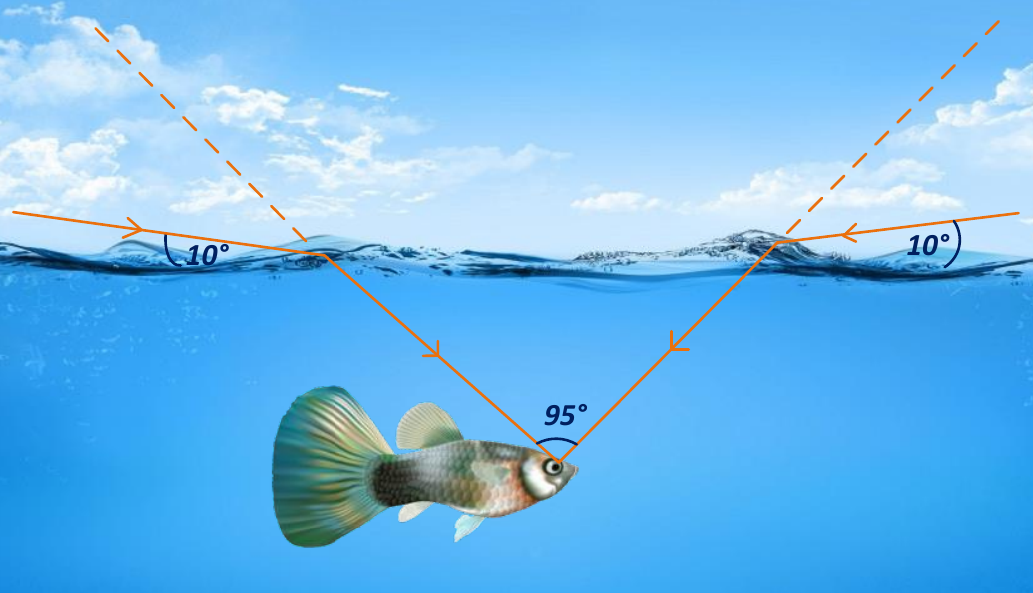}
\caption{\textbf{Illustration of fisheye perspective}. (top) Rays incident on a pinhole camera at wide angles cannot be imaged effectively beyond $60^\circ$. The addition of a fisheye lens dramatically increases the field-of-view to $190^\circ$ due to refraction. (bottom) Refraction of light rays at the surface of water causes compression of the horizon into a smaller field-of-view. }
\vspace{-0.4cm}
\label{fig:fisheyelens}
\vspace{-0.4cm}
\end{figure}

In theory, the field-of-view of a pinhole camera is $180^\circ$. However, in practice, due to the practical limitations of the size of the aperture and imager, it is not easy to get over $80^\circ$, as illustrated in Figure \ref{fig:fisheyelens} (top). Fisheye lenses are commonly used to effectively increase the field-of-view to $180^\circ$ or more. It is interesting to note that the term \textit{fisheye} is a bit of a misnomer, as illustrated in Figure \ref{fig:fisheyelens} (bottom). Due to the bending of light rays due to refraction at the junction of water and air surface, a large field-of-view of nearly $180^\circ$ is compressed to a smaller field-of-view of nearly $100^\circ$. A human swimmer would observe the same effect; it is nothing to do with the optics of fish's eye.\par

The development of fisheye cameras has a long history. Wood initially coined the term fisheye in 1908 and constructed a simple fisheye camera \cite{wood1908fisheye}, a fact that is acknowledged in the naming of the recently released \textit{WoodScape} dataset of automotive fisheye video \cite{yogamani2019woodscape}. This water-based lens was replaced with a hemispherical lens by Bond \cite{bond1922fisheye}, and thus began the optical development of fisheye cameras. Miyamoto \cite{miyamoto1964fisheye} provided early insight into the modelling of geometric distortion in fisheye cameras, suggesting the use of equidistant, stereographic, and equisolid models. These models were already known in the field of cartography (\eg, \cite{thomas1952projections} and many others).\par
% -------------------------------------------------
% -------------------------------------------------
\textbf{Applications:} Fisheye cameras offer a significantly wider field-of-view than standard cameras, often with a $180^\circ$ field-of-view or even greater. This can offer several advantages, not least that fewer cameras can be used to achieve complete coverage. The first successful commercial application of fisheye cameras was in photography, particularly in the entertainment industry, where the fisheye lens effect became a stylistic element. A video by Vox \cite{Voxfisheye} provides an excellent overview of the history of its usage. The second successful area of application is video surveillance, where the hemispherical lens surface can be commonly seen in modern surveillance systems \cite{kim2016fisheye}. More recently, wide-angle lens cameras are commonly used in virtual reality headsets \cite{defanti2009starcave}. They are also commonly used in underwater robotics \cite{meng2018underwater}, and aerial robotics \cite{qiu2017model}.\par

Automotive is one of the important application areas of fisheye cameras where more advanced visual perception is necessary. 
The first wide-angle rear-view camera and a TV display were deployed in General Motors' Buick Centurion concept model in 1956. In 2018, a rear-view fisheye camera was mandated in the United States to reduce accidents during reversing \cite{sunstein_2019}. In 2008, surround-view cameras were deployed by BMW for park view \cite{hughes2009wide}. Surround-view cameras have become a commonly used feature in many vehicles. They were subsequently used for computer vision applications like cross-traffic alerts \cite{Bandyopadhyay2020}, object detection \cite{rashed2021generalized}, and automated parking \cite{heimberger2017computer}. Figure \ref{fig:surroundview} (top) illustrates the position of the cameras and sample images of a surround-view system. Figure \ref{fig:surroundview} (bottom) shows the near-field region, and it forms the primary sensor for $360^\circ$ sensing around the vehicle. Surround visualization for the driver by stitching the four cameras is also illustrated within the smaller box. \\
% -------------------------------------------------
% -------------------------------------------------

\begin{figure}[t]
\centering
\captionsetup{singlelinecheck=false, font=footnotesize, belowskip=0pt}
\includegraphics[width=0.75\columnwidth,page=2]{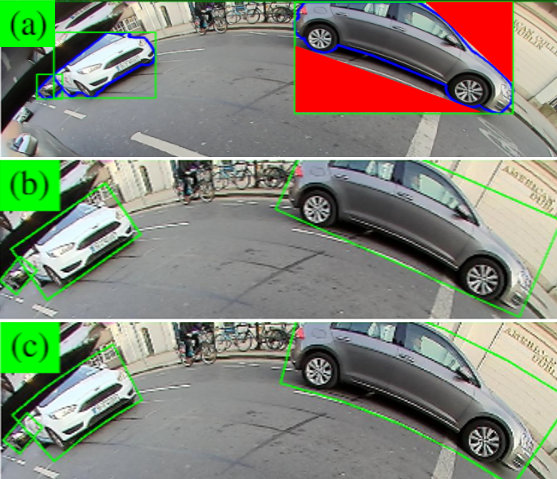}
\caption{\textbf{Standard bounding box is not a good object representation for fisheye images}. (a) Red pixels within the bounding box show a large area that does not contain the object. Oriented box (b) and curved bounding box (c) are better representations \cite{rashed2021generalized}. }
\label{fig:fisheyebox}
\vspace{-0.4cm}
\end{figure}

% -------------------------------------------------
Fisheye cameras have several challenges, however. The most obvious is that they exhibit a strong radial distortion that cannot be corrected without disadvantages, including reduced field-of-view and resampling distortion artifacts at the periphery~\cite{kumar2020unrectdepthnet}. Appearance variations of objects are larger due to the spatially variant distortion, particularly for close-by objects. This increases the learning complexity of a Convolutional Neural Network (CNN), which uses translation invariance as an inductive bias and increases sample complexity as the model must learn the appearance of all the distorted versions of an object. In addition, the commonly used application of object detection using a bounding box becomes more complex as the bounding box does not provide an optimal fit for fisheye distorted objects, as illustrated in Figure \ref{fig:fisheyebox}. More sophisticated representations instead of a rectangular box, such as a curved bounding box exploiting the known radial distortion of fisheye cameras, were explored in \cite{rashed2021generalized}.
Fisheye perception is a challenging task, and despite its prevalence, it is comparatively less explored than pinhole cameras.\par

In the case of cameras without significant fisheye distortion, there is a very common geometry associated with them, being the \textit{pinhole} model. One may first consider the intersection of a ray with a single planar surface at some fixed distance from the projection center. All models of the distortion due to the lens for such cameras then are simply designed to shift the intersection point position radially from the projection center on the plane. In a way, fisheye algorithm development has been complicated by the lack of a unifying geometry. Many models use different properties to describe fisheye projection. 
One of the aims of this paper is to examine common models and demonstrate that several of the models are highly related to one another. Several models can be seen as specific cases of the General Perspective Mapping or Ellipsoidal General Perspective Mapping, both of which have been known for many decades in other fields of science \cite{fitzgibbon2001divisionmodel}. We show that a few of the presented models are even re-derivations of existing models. 
Thus we attempt to map a path through the many proposed models and consider them in several classes. For example, we could consider a class of \textit{on-image} models, in which the fisheye projection is measured as a deviation from pinhole projection, \eg, \cite{basu1992pfet, devernay2001fovmodel}. Alternatively, we could consider a model in which the ray projection angle is manipulated at the projection center (\eg, \cite{yogamani2019woodscape, kannala2006fisheye}). Others still propose the use of a series of projections onto different surfaces to model fisheye distortion, for example, \cite{geyer2000ucm, khomutenko2016eucm, usenko2018doublesphere}, which we can refer to as \textit{spherical models}.
% -------------------------------------------------
% -------------------------------------------------

\textbf{Relation to other sensors:} Automated parking systems are typically designed using fisheye cameras and sonar \cite{heimberger2017computer}. Sonar is typically used in the front and rear, and it is very reliable to detect near-field obstacles accurately \cite{park2008parking}. However, its range is typically limited to about 5 meters. Additionally, the information is very sparse, making it impossible to obtain richer information about the scene. Typically, a classical late fusion approach combines the perception output of the fisheye camera, and sonar \cite{suhr2013sensor, heimberger2017computer}. More recently, an array of short-range radars (SRR) providing $360^\circ$ coverage, which is used for urban driving applications, are being reused for near-field sensing applications like parking. They are significantly denser than sonar and have a range of 30 meters. However, they do not cover the entire near-field, and there are some blind spots. Radar is additionally limited in that it cannot detect road markings and has limited performance in object classification \cite{tang2022}. Parking space detection using SRR is discussed in more detail in \cite{loeffler2015parking}. Fusion of fisheye camera and SRR is typically performed in a classical dynamic occupancy grid fusion framework \cite{schmid2011parking}. CNN-based fusion approaches are emerging as well \cite{lekic2019automotive}. LiDAR is a far-field sensor with a range of over 200 meters, and thus it is typically not combined with near-field fisheye cameras. Varga \etal \cite{varga2017super} have attempted to combine fisheye camera and LiDAR to provide a unified $360^\circ$ environmental model, but there are blind spots in the near field. Classification of objects in LiDAR has extremely limited performance \cite{tang2022}. To summarize, other near-field sensors like radar and sonar capture limited information about the scene, and thus they cannot operate independently to perform near-field perception.

This paper is intended to be a broad overview and survey complementing our previous work \cite{eising2021near} which is relatively a narrow discussion of our concrete architecture and implementation of surround-view perception. We list few other review papers which are related to our paper. In \cite{heimberger2017computer}, a brief survey of computer vision for the specific use case of automated parking was provided. In \cite{Hughes2009WideangleCT}, an early survey is provided on surround-view monitoring, though no perception tasks are discussed. Finally, \cite{malik2016_3rs} provides a comprehensive review of vision tasks, but not specifically for automotive surround-view systems.

The paper is organized as follows. In Section \ref{sec:cameramodels}, we discuss some of the commonly used models and build a taxonomy of these methods establishing equivalences and specialization.
In Section \ref{sec:svs}, we introduce the automotive setup of four fisheye cameras forming a near-field surround view system and discuss basic constructs like calibration, rectification, and geometric primitives. Section \ref{sec:tasks} covers a detailed survey of visual perception tasks on surround-view cameras. Section \ref{sec:challenges} discusses future research directions to be explored by the community. Section \ref{sec:conclusion} provides concluding remarks. 
\par
% -------------------------------------------------

% Related work 
% -------------------------------------------------
\section{Fisheye Camera Models}
\label{sec:cameramodels}

In this section, we provide a survey of several of the more popular fisheye camera models. The aim is to provide a comprehensive list of possible models using a unified notation.  For a developer, this could be seen as a tool to guide the choice of model for a given application. One could attempt to use the simpler, more specialized models, and, depending on the specific application, extend the development to one of the more general models in the case that errors remain high for a given camera following calibration.\par

\textbf{Notation and Terminology:}
Matrices are denoted by $\matr{A} \in \spce{R}^{m \times n}$. The usual notation for ordinary vectors $\vec{v} \in \spce{R}^n$ will be used, represented as $n$-tuples. Specifically, points in $\spce{R}^3$ will be denoted as $\vec{X}~=~(X,Y,Z)^\T$, and a point in the set of image points $I^2$ is denoted as $\vec{u}=(u, v)^\T$.
The unit sphere is defined by ${S}^2 = \left\{\vec{s}\in\spce{R}^3 \, | \, \|\vec{s}\|=1\right\}$, and points on the unit sphere are represented as 3-vectors, i.e., $\vec{s}=(x,y,z)^\T$.

We can define a mapping from $C^3 \subseteq \spce{R}^3$ to the image as
$$\prjt : C^3 \rightarrow I^2$$
where $C^3$ denotes the set of points for which the projection $\pi$ is defined. $I^2 \subseteq \spce{R}^2$ denotes the image following projection from $C^3$. 
$\theta$ (usually in radians) is used to denote the field-angle (angle against the $Z$-axis) of the imaged point, and $\theta_{max}$ indicates the maximum field-angle of the model.

\begin{figure}
  \centering
   \captionsetup{font=footnotesize, belowskip=0pt}
  \includegraphics[width=\linewidth]{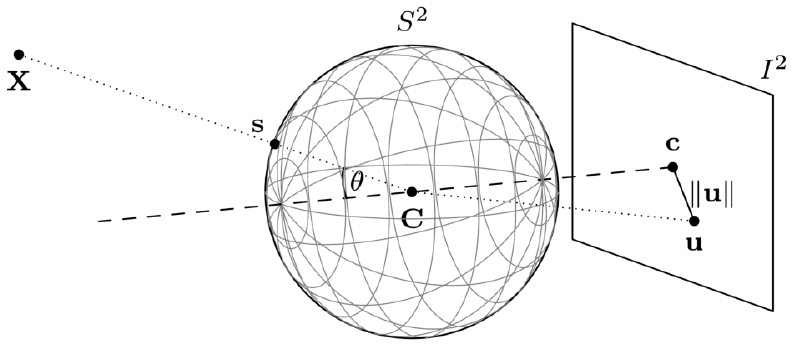}
  \caption{\bf Relationship between fisheye image point $\mathbf{u}$ and and its equivalent point $\mathbf{s}$ on the unit sphere, with $\mathbf{s}$ lying on the same ray as $\mathbf{X}$.}
  \label{fig:fisheye}
  \vspace{-0.4cm}
\end{figure}

A true inverse of $\pi$ is naturally not possible. However, we can define an unprojection function mapping from the image domain to the unit central projective sphere
$$\prjt^{-1}: {I}^2 \rightarrow {S}^2$$
In some cases, the analytic unprojection $\prjt^{-1}(\vec{u})$ does not exist or has singularities. Figure \ref{fig:fisheye} demonstrates the relationship between the image points and the unit sphere.

We also use $\|\vec{u}\| = \radl(\theta)$ to denote the radial form of the projection function. That is, this is a function that maps the field angle to a radial distance on the image plane (from the distortion center). The radial unprojection function is denoted $\theta = \radl^{-1}(\|\vec{u}\|)$. The radial to incident angle unprojection is a true inverse, unlike the unprojection to the image sphere. Occasionally, we will have the need to refer to two image points, a distorted and an undistorted point. In this case, we will use the subscript $d$ and $u$ to distinguish (\eg, $\vec{u}_d$ and $\vec{u}_u$). On-image mappings radially warp an image from its distorted point to the undistorted point (\ie, from $\|\vec{u}_d\|$ to $\|\vec{u}_u\|$) on the image. We denote this mapping as $\|\vec{u}_u\| = \onimg(\|\vec{u}_d\|)$, and its inverse $\|\vec{u}_d\| = \onimg^{-1}(\|\vec{u}_u\|)$.

When discussing the models below, we use subscripts to denote the parameters and functions for each of the different models. Specifically, we use subscript $p$ for the pinhole model, $e$ for the equidistant, $s$ for the stereographic, $o$ for the orthographic, $eo$ for the extended orthographic, $div$ for division, $fov$ for field-of-view, $ucm$ for Unified Camera Model and $ds$ for Double-Sphere.

\textbf{Pinhole Camera Model:}
The pinhole camera model is the standard projection function used in many areas of computer vision and robotics when the research is limited to considering standard field-of-view cameras. The pinhole model is given by
\begin{equation} \label{eqn:pinhole}
    \prjt_p(\vec{X}) = \frac{f}{Z} \mat{X\\ Y}
\end{equation}
or, if we consider it as a radial function
\begin{equation} \label{eqn:pinholeradial}
    \radl_p(\theta) = f\tan{\theta}
\end{equation}
where $\theta$ is the field angle of the projected ray. Note that the parameter $f$ is sometimes referred to as the focal length.

The unprojection functions are
\begin{align}
\radl^{-1}_p(\|\vec{u}\|) & = \atan\left(\frac{\|\vec{u}\|}{f}\right) , \quad
\prjt^{-1}_p(\vec{u}) & = \frac{(u, v, f)^\T}{\|(u, v, f)^\T\|}
\end{align}
The pinhole model is defined for the set of points $C^3 = \left\{\vec{X}~\in~\spce{R}^3~|~Z~>~0\right\}$. The points map to the entire image plane, \ie, $I^2 = \spce{R}^2$, and $\theta_{max} = \pi/2$. In practice, however, even when radial distortion is considered, the pinhole model is rarely of use for points with field-angle $\theta > 60^\circ$. \par

% -------------------------------------------------
\begin{figure*}
 \captionsetup{font=footnotesize}
     \centering
     \begin{subfigure}[b]{0.24\textwidth}
         \centering
         \includegraphics[width=\textwidth]{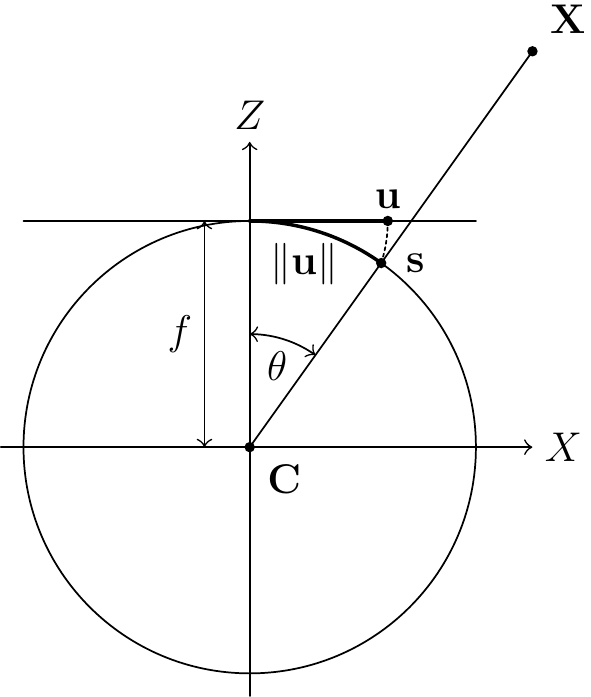}
         \caption{Equidistant}
         \label{fig:equidistant}
     \end{subfigure}
     \hfill
     \begin{subfigure}[b]{0.24\textwidth}
         \centering
         \includegraphics[width=\textwidth]{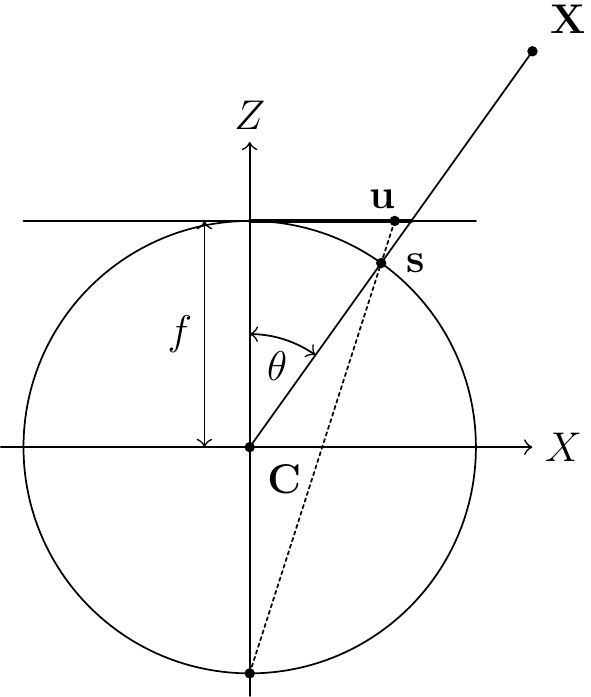}
         \caption{Stereographic}
         \label{fig:stereographic}
     \end{subfigure}
     \hfill
     \begin{subfigure}[b]{0.24\textwidth}
         \centering
         \includegraphics[width=\textwidth]{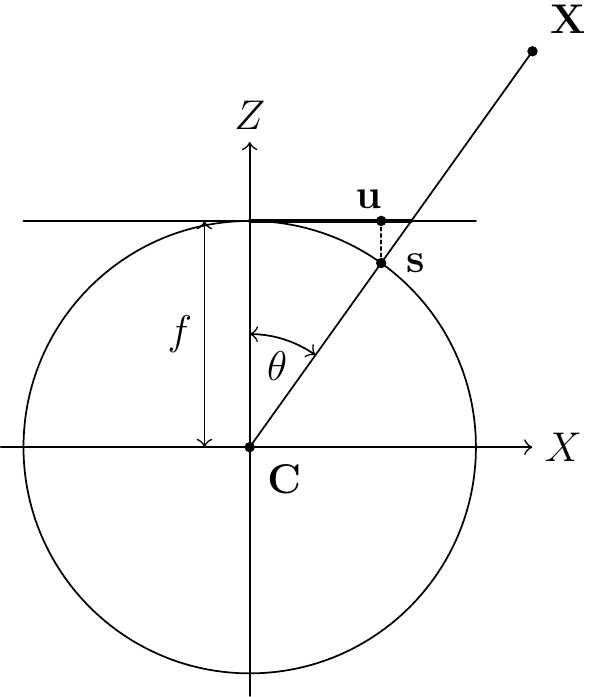}
         \caption{Orthographic}
         \label{fig:orthographic}
     \end{subfigure}
     \hfill
     \begin{subfigure}[b]{0.24\textwidth}
         \centering
         \includegraphics[width=\textwidth]{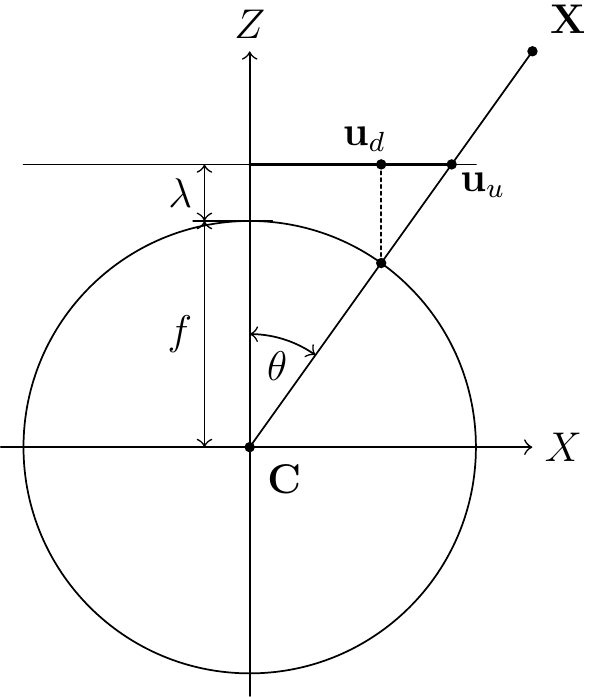}
         \caption{Extended Orthographic}
         \label{fig:extended_orthographic}
     \end{subfigure}
        \caption{\bf Classical Geometric models.}
        \label{fig:classical}
    \vspace{-0.4cm}
\end{figure*}
% -------------------------------------------------
% -------------------------------------------------
\subsection{Classical Geometric Models}
We refer to the models discussed in this section as {\em classical}, as they have been researched for at least six decades \cite{miyamoto1964fisheye}.
% -------------------------------------------------
\textbf{Equidistant Projection:}
In the \textit{equidistant fisheye model}, the projected radius $\radl_e(\theta)$ is related to the field angle $\theta$ through the simple scaling by the equidistant parameter $f$ (see Figure \ref{fig:equidistant}). That is
\begin{equation} \label{eqn:equidistantradial}
    \radl_e(\theta) = f\theta , \quad \prjt_e(\vec{X}) = \frac{f\theta}{d} \mat{X\\Y}
\end{equation}
where
\begin{align} \label{eqn:equidistant}
    d = \sqrt{X^2 + Y^2}  , \quad
    \theta = \acos \left(\frac{Z}{\|\vec{X}\|}\right)
\end{align}
The unprojection functions are
\begin{align}
\radl^{-1}_e(\|\vec{u}\|) & = \frac{\|\vec{u}\|}{f} , \quad
\prjt^{-1}_e(\vec{u}) & = \mat{\frac{u}{\|\vec{u}\|}\cdot\sin\left(\frac{\|\vec{u}\|}{f}\right)\\\frac{v}{\|\vec{u}\|}\cdot\sin\left(\frac{\|\vec{u}\|}{f}\right)\\\cos\left(\frac{\|\vec{u}\|}{f}\right)}
\end{align}
The equidistant projection is valid for the points $C^3~=~\spce{R}^3~\setminus~(0,0,0)^\T$, $I^2 = \left\{\vec{u}~\in~\spce{R}^2~|~\|\vec{u}\|~\leq~f\pi\right\}$, and $\theta_{max} = \pi$.\par

\textbf{Stereographic Projection:}
As with the equidistant model, in \textit{stereographic projection}, the center of projection of $\vec{X}$ to the projection sphere is $\vec{C}$ (Figure \ref{fig:stereographic}).
The stereographic projection is therefore described by
\begin{equation} \label{eqn:stereographicradial}
    \radl_s(\theta) = 2\, f\, \tan\left(\frac{\theta}{2}\right) , \quad
    \prjt_s(\vec{X}) = \frac{2\, f}{Z + \|\vec{X}\|} \mat{X\\Y}
\end{equation}
The unprojection functions, which we shall need later, are
\begin{align} \label{eqn:stereographicradialinv}
    \radl^{-1}(\|\vec{u}\|) & = 2\, \atan\left(\frac{\|\vec{u}\|}{2 f}\right) \\
    \prjt^{-1}_s(\vec{u}) & = \frac{1}{4f^2 + \|\vec{u}\|^2} \mat{4fu\\4fv\\4f^2 - \|\vec{u}\|^2}
\end{align}
The stereographic projection is valid for the points $C^3~=~\spce{R}^3~\setminus~(0,0,0)^\T$, and maps these points to the entire image plane, \ie, $I^2 = \spce{R}^2$. As such, the maximum field-angle is $\theta_{max} = \pi$.\par
% -------------------------------------------------
\textbf{Orthographic Projection:}
Similar to the previous projections models, the \textit{orthographic projection} begins with a projection to the sphere (Figure \ref{fig:orthographic}). This is followed by an orthogonal projection to the plane. The orthographic projection is therefore described by
% -------------------------------------------------
\begin{equation} \label{eqn:orthographicradial}
    \radl_o(\theta)= f \sin\theta , \quad
    \prjt_o(\vec{X}) = \frac{f}{\|\vec{X}\|} \mat{X\\Y}
\end{equation}
The unprojection functions are
\begin{align} \label{eqn:orthographicradialinv}
    \radl^{-1}_o(\|\vec{u}\|) & = \asin\left(\frac{\|\vec{u}\|}{f}\right) \nonumber \\
    \prjt^{-1}_o(\vec{u}) & = \frac{1}{f} \mat{u\\v\\\sqrt{f_o^2 - \|\vec{u}\|^2}}
\end{align}
Here, $I^2=\spce{R}^2$,  $C^3~=~\left\{\vec{X}~\in~\spce{R}^3~|~Z~>~0\right\}$, and $\theta_{max}~=~\pi/2$. These unprojection functions are well defined, as $f \geq \|\vec{u}\|$, which is enforced by the original projection (\ref{eqn:orthographicradial}).\par
% -------------------------------------------------
\textbf{Extended Orthographic Model:}
The \textit{Extended Orthographic Model} \cite{kim2014model}, as demonstrated by Figure \ref{fig:extended_orthographic}, extends the classical orthographic model by freeing the projection plane from being tangential to the projection sphere, allowing an offset $\lambda$.
The distorted projection remains the same as equations (\ref{eqn:orthographicradial}). 
However, the relationship between the distorted and undistorted radial distances and its inverse is given by
% -------------------------------------------------
\begin{align} 
    \onimg_{eom}(\|\vec{u}_d\|) & = \frac{(\lambda + f)\|\vec{u}_d\|}{\sqrt{f^2 - \|\vec{u}_d\|^2}} \\
    \onimg_{eom}^{-1}(\|\vec{u}_u\|) & = \frac{f\|\vec{u}_d\|}{\sqrt{(\lambda + f)^2 + \|\vec{u}_u\|^2}}
\end{align}
% -------------------------------------------------
This is slightly simplified representation to that presented in \cite{kim2014model}, and assumes that $f$ and $(\lambda + f)$ are positive, which is entirely practical constraints.
The extended orthographic model has the same domain and co-domain as the standard orthographic model.

\textbf{Extended Equidistant Model:} In fact, the extended orthographic model is simply a conversion from a projection to an on-image map. Many models can be converted to on-image mappings in the same manner as the extended orthographic model. We give just one example of the equidistant model. 

Rearranging (\ref{eqn:equidistantradial}) such that $\theta = \|\vec{u}_s\| / f$, substituting into (\ref{eqn:pinholeradial}), and letting the focal length of (\ref{eqn:pinholeradial}) be $f + \lambda$, we get the on-image mapping for the equidistant model. Following similar steps, we can also obtain the inverse.
\begin{align}
    \onimg_e(\|\vec{u}_d\|) & = (f + \lambda) \tan\left(\frac{\|\vec{u}_d\|}{f}\right) \\
    \onimg^{-1}_e(\|\vec{u}_u\|) & = f \atan\left(\frac{\|\vec{u}_u\|}{f + \lambda}\right) 
\end{align}
This is described in \cite{hughes2010equidistant}, albeit without the additional scaling parameter $\lambda$. We could follow the same steps above to obtain an \textit{Extended Stereographic Model} as well. $C^3$, $I^2$ and $\theta_{max}$ for these extended models are the same as for the orthographic model. 

\subsection{Algebraic models}

We provide a short discussion on algebraic models of fisheye cameras, specifically polynomial models, and the division model. The polynomial model discussion we provide for completeness, though we concentrate on the geometric models for the remainder of the paper.\par 
% -------------------------------------------------
\textbf{Polynomial Models:} \label{sec:polynomialmodel}
The classical \textit{Brown–Conrady model} of distortion for non-fisheye cameras \cite{brown1966, conrady1919decentred} uses an odd-termed polynomial, $\|\vec{u}_d\| = P_n(\|\vec{u}_u\|)$, to describe the radial distortion on the image (\ie mapping $\|\vec{u}_u\|$ to $\|\vec{u}_d\|$), where $P_n$ represents some arbitrary $n$th order polynomial. Despite its age, the Brown-Conrady model is the standard distortion model in software implementations for non-fisheye cameras \cite{opencv_library, MATLAB:2021}.
To account for fisheye distortion, an on-image polynomial model known as the \textit{Polynomial Fisheye Transform} (PFET), was proposed in \cite{basu1992pfet}. The difference between the PFET and the Brown-Conrady model is that the PFET allows both odd and even exponents to account for the added distortion encountered in fisheye cameras.

A class of polynomial fisheye models exist, in which the mapping of the field angle to the image plane is via a polynomial, i.e $\radl_P(\theta) = P_n(\theta)$, using the angle of incidence instead of the undistorted radius.
For example, Kannala-Brandt \cite{kannala2006fisheye} (and as implemented in the popular \textit{OpenCV} software \cite{opencv_library}) propose an polynomial model of order $n=5$, or more, with odd exponents only.
In \cite{yogamani2019woodscape}, an $n=4$ polynomial containing both even and odd exponents is proposed. Neither model used a constant coefficient term in the polynomial, as doing so would lead to an undefined area in the center of the image. In \cite{ying2006poly} a fifth order polynomial is proposed, but they reduce it to four independent parameters if the fisheye radius and the field-of-view are known. 
All the above could be interpreted as generalization of the equidistant model, which is a first order polynomial. In this case, the projection sphere is replaced by some surface defined by the given polynomial. However, this is forcing a geometric interpretation with little utility.

The \textit{MATLAB Computer Vision Toolbox} \cite{matlab2020calibration} and the NVidias DriveWorks SDK \cite{nvidia2020calibration} include implementations of a polynomial-based fisheye model provided in \cite{Scaramuzza2006Calibration}. In this case, polynomials are used to model both the projection and unprojection, negating the need for a numerical approach to invert a projection (which is a major computational problem for polynomial-based models).  
Note, that both polynomials are not the inverse of each other, but two different functions. These polynomials are calibrated independently, which can make it unusable if iterative approaches that project and unproject points for several times.\par
% -------------------------------------------------
\textbf{Division Model:}
The \textit{division model} \cite{fitzgibbon2001divisionmodel} of radial distortion gained some popularity due to the nice property that, at least for the single parameter variant, straight lines project to circles in the image \cite{Wildenauer2013DivisionModel, antunes2017unsupervised, bukhari2013automatic}
, and for many lenses, the single parameter variant performs very well \cite{Courbon2012FisheyeAnalysis}. The model and its inverse are given by
% -------------------------------------------------
\begin{align} \label{eqn:divisionmodelradialund}
    \onimg_{div}(\|\vec{u}_d\|) & = \frac{\|\vec{u}_d\|}{1-a\|\vec{u}_d\|^2} \\
    \onimg_{div}^{-1}(\|\vec{u}_u\|) & = \frac{\sqrt{1 + 4 a \|\vec{u}_u\|^2} - 1}{2 a \|\vec{u}_u\|}
\end{align}
% -------------------------------------------------
This was extended in \cite{hughes2010equidistant} by adding an additional scaling parameter, which improved the modeling performance for certain types of fisheye lens. While the division model was originally presented as an \textit{on-image} mapping, it can be expressed as the projection function
\begin{align}  \label{eqn:divisionmodel}
    \radl_{div}(\theta) &= \frac{\sqrt{1 + 4 a f \tan^2\theta} - 1}{2 a f \tan\theta} \nonumber \\
    \prjt_{div}(\vec{X})  &= \frac{f r'_d}{Z r'_u} \mat{X\\Y} \nonumber \\
    r_u' &= \frac{\sqrt{X^2 + Y^2}}{Z} , \quad
    r_d' = \frac{\sqrt{1 + 4br_u'^2} - 1}{2 b r_u'}
\end{align}
% -------------------------------------------------
The radial projection function $\radl_{div}(\theta)$ is simply obtained by substituting the pinhole model (\ref{eqn:pinholeradial}) into (\ref{eqn:divisionmodelradialund}). $f$, in this case, can be thought of as the parameter of the pinhole model once distortion has been addressed by the division model.
% -------------------------------------------------
The unprojection of the division model is
\begin{align}
    \radl^{-1}_{div}(\|\vec{u}\|) &= \atan2\left({\|\vec{u}\|, f\left(1 - a\|\vec{u}\|^2\right)}\right) \nonumber \\
    \prjt^{-1}_{div}(\vec{u}) &= \frac{(u', v', f)^\T}{\|(u', v', f)^\T\|}  \nonumber , \quad
    \vec{u}' = \frac{1}{1-a\|\vec{u}\|^2}\vec{u}
\end{align}
The projection functions and the on-image mapping have the same domain, $C^3~=~\left\{\vec{X}~\in~\spce{R}^3~|~Z~>~0\right\}$, $I^2 = \spce{R}^2$, and $\theta_{max} = \pi/2$.
% -------------------------------------------------
\subsection{Spherical models}

A set of more recent (at least, from the last couple of decades) fisheye models are also considered, based on the projection of the point to a unit sphere (or its affine generalisation). \par
% -------------------------------------------------
\textbf{Field-of-View Model:}
The field-of-view model \cite{devernay2001fovmodel} and its inverse is defined by
\begin{align}  \label{eqn:fovmodelradial}
\onimg_{fov}(\|\vec{u}_d\|) & = \frac{\tan(\|\vec{u}_d\| \omega)}{2\tan\frac{\omega}{2}} \\
\onimg_{fov}^{-1}(\|\vec{u}_u\|) & = \frac{\atan \left(2\|\vec{u}_u\|\tan\frac{\omega}{2}\right)}{\omega}
\end{align}
The parameter $\omega$ approximates the camera field-of-view, though not exactly \cite{devernay2001fovmodel}.
% -------------------------------------------------
This is an \textit{on-image} model, like the Division Model, where $\|\vec{u}_u\|$ and $\|\vec{u}_d\|$ define undistorted and distorted radii on the image plane. Alternatively, it can be expressed as a projection function \cite{usenko2018doublesphere}.
\begin{align}  \label{eqn:fovmodel}
    \radl_{fov}(\theta) &= \frac{\atan\left(2f\tan\theta\tan\frac{\omega}{2}\right)}{\omega} \\ 
    \prjt_{fov}(\vec{X}) &= \frac{f r_d'}{r_u'} \mat{X\\Y} \nonumber \\
    r_u' &= \sqrt{X^2 + Y^2} ,\quad
    r_d' = \frac{\atan2(2 r_u' \tan(\frac{\omega'}{2}), Z)}{\omega'} \nonumber
\end{align}
The unprojection is given by
\begin{align}  \label{eqn:fovmodel_inv}
    \radl^{-1}_{fov}(\|\vec{u}\|) &= \atan \left(\frac{\tan\left(\omega' \|\vec{u}\|\right)}{2 f \tan \frac{\omega'}{2}}\right) \\
    \prjt^{-1}_{fov}(\vec{u}) &= \mat{
       \frac{u}{f} \cdot \frac{sin(r_d'\omega')}{2r_d'\tan(\frac{w}{2})} \\
       \frac{v}{f} \cdot \frac{sin(r_d'\omega')}{2r_d'\tan(\frac{w}{2})} \\
       \cos(r_d'\omega')
    } , \quad
    r_d' = \frac{\|\vec{u}\|}{f}
\end{align}
A nice artefact of expressing the field-of-view model as a projection function is that the domain of the projection $\prjt_{fov}(\vec{X})$ covers all of $C^3 = \spce{R}^3 \setminus (0,0,0)^\T$. In contrast, the on-image mapping form of the field-of-view model is restricted to mapping image points where $C^3~=~\left\{\vec{X} \in \spce{R}^3 \, | \, Z > 0\right\}$, which is true of any on-image mapping. The set of imaged points is $I^2 = \left\{\vec{u}~\in~\spce{R}^2~|~\|\vec{u}\|~\leq~\frac{\pi}{\omega'}\right\}$, and $\theta_{max} = \pi$. We shall soon show that the field-of-view model is the equivalent of the equidistant model, and as such is a spherical projection.\par
\begin{figure}
     \captionsetup{font=footnotesize, belowskip=0pt}
     \centering
     \begin{subfigure}[b]{0.32\columnwidth}
         \centering
         \includegraphics[width=\textwidth]{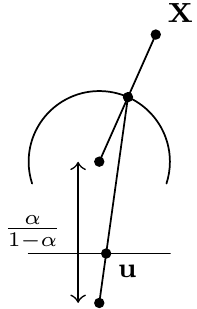}
         \caption{UCM}
         \label{fig:geo_ucm}
     \end{subfigure}
     \hfill
     \begin{subfigure}[b]{0.32\columnwidth}
         \centering
         \includegraphics[width=\textwidth]{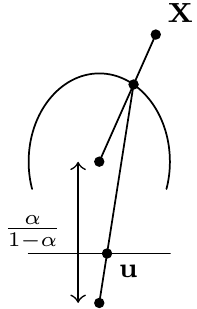}
         \caption{E-UCM}
         \label{fig:geo_eucm}
     \end{subfigure}
     \hfill
     \begin{subfigure}[b]{0.32\columnwidth}
         \centering
         \includegraphics[width=\textwidth]{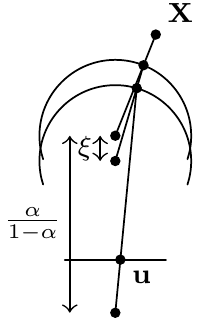}
         \caption{Double Sphere}
         \label{fig:geo_ds}
     \end{subfigure}
        \caption{\textbf{Spherical models.} The UCM (a) consists first of a projection to the unit sphere, followed by a perspective projection. The E-UCM replaced the sphere with an ellipsoid, with coefficient $\beta$. The DS model added a second unit sphere projection to the UCM, with the distance between the spheres being $\xi$.}
        \label{fig:spherical_models}
        \vspace{-0.6cm}
\end{figure}

% -------------------------------------------------
\textbf{Unified Camera Model:}\label{sec:ucm}
The UCM was initially used to model catadioptric cameras \cite{geyer2000ucm}, and later was shown to be useful when modelling fisheye cameras \cite{ying2004division, courbon2007generic}. It has been shown to perform well across a range of lenses \cite{Courbon2012FisheyeAnalysis}. First, the point $\vec{X}$ is projected to a unit sphere, followed by a projection to a modeled pinhole camera (Figure \ref{fig:geo_ucm}). We present the version with better numerical properties from \cite{usenko2018doublesphere}.
\begin{align} \label{eqn:ucm}
    \radl_{ucm}(\theta) &= \frac{f\sin\theta}{(1-\alpha)\cos\theta + \alpha} \\
    \prjt_{ucm}(\vec{X}) &= \frac{f}{\alpha\|\vec{X}\| + (1-\alpha)Z} \mat{X\\Y}
\end{align}
% -------------------------------------------------
The unprojection of the UCM is given by
\begin{align} \label{eqn:ucm_inv}
    \prjt_{ucm}^{-1}(\vec{u}) &= K \mat{u\\v\\\frac{f}{1-\alpha}} - \mat{0\\0\\\frac{\alpha}{1 - \alpha}} \\
    K &=  \frac{\alpha f + (1 - \alpha)\sqrt{(1-2\alpha)\|\vec{u}\|^2 + f^2}}{(1 - \alpha)^2\|\vec{u}\|^2 + f^2}
\end{align}
$\radl^{-1}_{ucm}(\theta)$ is a complicated equation (more so than the above) and as such is not shown here. The domain of the projection and the radial function is given as
\begin{align}
    & C^3 = \begin{cases}
    \{\vec{X}\in\spce{R}^3 \, | \, Z>\|\vec{X}\|\frac{\alpha}{\alpha - 1}\}, & \text{if }\alpha \le 0.5\\
    \{\vec{X}\in\spce{R}^3 \, | \, Z>\|\vec{X}\|\frac{\alpha - 1}{\alpha}\}, & \text{if }\alpha > 0.5
    \end{cases} \\
    & \theta_{max} = \pi - \acos\left(\frac{1-\alpha}{\alpha}\right) \\
    & I^2 = \begin{cases}
    \spce{R}^2, & \text{if }\alpha \le 0.5\\
    \{\vec{u}\in\spce{R}^2 \, | \, \|\vec{u}\|^2\le\frac{f^2}{2\alpha - 1}\}, & \text{if }\alpha > 0.5
    \end{cases} 
\end{align}
When $\alpha < 0.5$, the pinhole projection point is inside the unit sphere, outside when $\alpha > 0.5$, and on the sphere when $\alpha = 0.5$.

% -------------------------------------------------
% -------------------------------------------------
\textbf{Enhanced Unified Camera Model:}\label{sec:eucm}
The UCM was extended by the Enhanced UCM \cite{khomutenko2016eucm} (Figure \ref{fig:geo_eucm}), which generalizes the spherical projection with a projection to an ellipsoid (or, in fact, a general quadratic surface), and was able to demonstrate some accuracy gain. The E-UCM is given by
\begin{align} \label{eqn:eucm}
    \radl_{eucm}(\theta) &= \frac{f\sin\theta}{\alpha\sqrt{\beta\sin^2\theta + \cos^2\theta} + (1 - \alpha)\cos\theta}\\
    \prjt_{eucm}(\vec{X}) &= \frac{f}{\alpha d + (1 - \alpha) Z} \mat{X\\Y}
\end{align}
where $d = \sqrt{\beta(X^2 + Y^2) + Z^2}$, and $\beta$ is the ellipse coefficient. The unprojection function is not pretty for the EUCM, and the readers are referred instead to \cite{khomutenko2016eucm}. The set of valid points and angles is
\begin{align}
    & C^3 = \begin{cases}
    \{\vec{X}\in\spce{R}^3 \, | \, Z>\|\vec{X}\|\frac{\beta\alpha}{\alpha - 1}\}, & \text{if }\alpha \le 0.5\\
    \{\vec{X}\in\spce{R}^3 \, | \, Z>\|\vec{X}\|\frac{\alpha - 1}{\beta\alpha}\}, & \text{if }\alpha > 0.5
    \end{cases} \\
    & \theta_{max} = \pi - \acos\left(\frac{1-\alpha}{\beta\alpha}\right) \\
    & I^2 = \begin{cases}
    \spce{R}^2, & \text{if }\alpha \le 0.5\\
    \{\vec{u}\in\spce{R}^2 \, | \, \|\vec{u}\|^2\le\frac{f^2}{\beta(2\alpha - 1)}\}, & \text{if }\alpha > 0.5
    \end{cases} 
\end{align}
% -------------------------------------------------
\textbf{Double-Sphere Model:}
Later still, the UCM was extended again by the double-sphere (DS) model \cite{usenko2018doublesphere}, which added a second unit sphere projection to enable more complex modeling (Figure \ref{fig:geo_ds}).
\begin{align} \label{eqn:ds}
    \radl_{ds}(\theta) &= \frac{f\sin\theta}{\sqrt{1+\xi^2 + 2\xi\cos\theta}\left(\frac{\alpha}{1 - \alpha} - \xi\right) + \left(\xi + \cos\theta\right)} \nonumber \\
    \prjt_{ds}(\vec{X}) &= \frac{f}{\alpha d_2 + (1 - \alpha)(\xi d_1 + Z)}\mat{X\\Y}\\
    d_1 &= \sqrt{x^2 + y^2 + z^2} \nonumber, \quad
    d_2 = \sqrt{x^2 + y^2 + (\xi d_1 + Z)^2} \nonumber
\end{align}
Convincing results are presented in \cite{usenko2018doublesphere} to demonstrate the effectiveness of the double-sphere model. The unprojection functions of this model are
\begin{align} \label{eqn:ds_unproj}
    \radl_{ds}^{-1}(\|\vec{u}\|) &= \asin\left(K\|\vec{u}\|\right) \nonumber \\
    \prjt^{-1}_{ds}(\vec{u}) &= K \mat{u\\v\\\zeta} \\
    K &= \frac{\zeta + \sqrt{\zeta^2 + (1 - \xi)^2\|\vec{u}\|^2}}{\zeta^2 + \|\vec{u}\|^2} \nonumber \\
    \zeta &= \frac{f - \alpha\|\vec{u}\|^2}{\alpha\sqrt{1 - (2\alpha - 1)\frac{\|\vec{u}\|^2}{f^2}} + 1 - \alpha} \nonumber
\end{align}
The valid ranges for projection and unprojection are
\begin{align}
    C^3 & = \{\vec{X}\in\spce{R}^3 \, | \, Z>-w_2\|\vec{X}\|\}\\
    \theta_{max} &=  \acos(-w_2) \nonumber \\
    I^2 & = \begin{cases}
        \spce{R}^2, & \text{if }\alpha \le 0.5\\
        \{\vec{u} \in \spce{R}^2 \, | \, \|\vec{u}\|^2 \le \frac{f^2}{2\alpha - 1}\}, & \text{if }\alpha > 0.5\\
    \end{cases} \nonumber \\
    w_2 & = \frac{w_1 + \xi}{\sqrt{2 w_1 \xi + \xi^2 + 1}}, \quad
    w_1 = \begin{cases}
        \frac{\alpha}{1-\alpha}, & \text{if }\alpha \le 0.5\\
        \frac{1-\alpha}{\alpha}, & \text{if }\alpha > 0.5
    \end{cases} \nonumber
\end{align}
% -------------------------------------------------
\subsection{Other Models}
While we have discussed many of the more popular fisheye projection models, still, this is not exhaustive. We have omitted the details of some models that would seem a little less popular, for whatever reasons. For example, Bakstein and Pajdla \cite{bakstein2002panoramic} proposed two extensions to the classical models.
A logarithm-based \textit{Fisheye Transform} (FET) was also proposed in \cite{basu1992pfet}, though the accuracy was low compared to other models. The hyperbolic sin-based model proposed in \cite{pervs2002nonparametric}, and later used for wide-angle cameras \cite{klanvcar2004wide}, is not discussed here, nor is the cascaded one-parameter division model \cite{mei2015radial}.\par
% -------------------------------------------------
\subsection{Unified Usage of Camera Models} \label{sec:commonality}
With the proliferation of fisheye models, it is natural to wonder if there is a commonality between some of the models, or even if there has been repetition in development of the models.\par
% -------------------------------------------------
\textbf{General Perspective Projection and Fisheye Models:}\label{sec:general_projection}
The unified camera model is in a class of general vertical perspective projections of a sphere, which is known in the fields of geodesy and cartography \cite{laubscher1965map, snyder1987maps}, with the addition of the trivial step of central projection to the spherical surface. The stereographic and the orthographic projections belong to this class as well. The stereographic projection has the pinhole projection center on the surface of the sphere, while the orthographic projection has an infinite focal length (hence the term orthographic). The link between the stereographic projection and the UCM is in fact described in \cite{geyer2000ucm}.\par

% -------------------------------------------------
\begin{figure}
 \captionsetup{font=footnotesize, belowskip=0pt}
  \centering
  \includegraphics[width=0.7\linewidth]{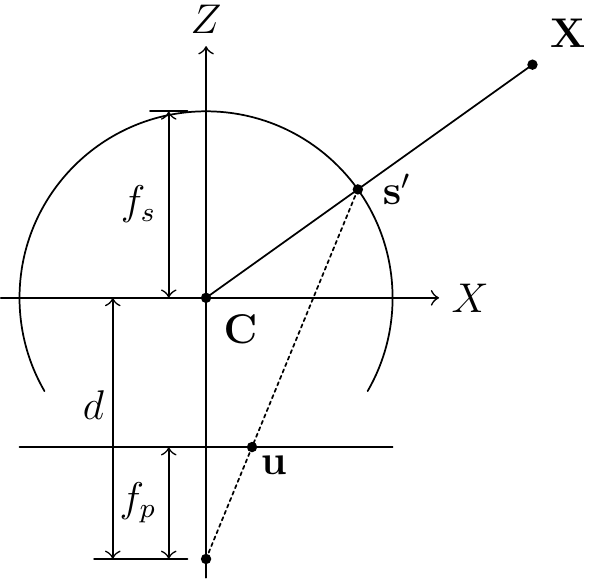}
  \caption{\textbf{The general perspective mapping is defined by a projection of a point to the sphere of radius $f_s$, followed by a perspective projection to pinhole model with focal length $f_p$.} The two projection centers are offset by $d$. As with the classical models, it is known in cartography for many decades \cite{laubscher1965map}.}
  \label{fig:generalperspective}
  \vspace{-0.4cm}
\end{figure}
% -------------------------------------------------

Let us begin by examining the general vertical perspective projection, described by Figure \ref{fig:generalperspective}. The pinhole camera is offset along the $Z$-axis by a distance $d$. The projection to the sphere is given by
% -------------------------------------------------
\begin{equation}
    \vec{s}' = f_s \frac{\vec{X}}{\|\vec{X}\|}
\end{equation}
% -------------------------------------------------
Here we use $\vec{s}' = (x', y', z')^\T$ for the point on the sphere of radius $f_s$, so as to distinguish it from $\vec{s}$ used previously to denote a point on the unit sphere. The point $\vec{u}$ is the pinhole projection of $\vec{s}'$
% -------------------------------------------------
\begin{equation} \label{eqn:generalperspective}
    \prjt(\vec{X}) = \frac{f_p}{z' + d} \mat{x'\\y'}
    =
     \frac{f_p}{Z + \frac{d}{f_s}\|\vec{X}\|} \mat{X\\Y}
\end{equation}
% -------------------------------------------------
The $+d$ translates the point $\vec{s}'$ from the sphere to the pinhole coordinate system. Thus, with the two parameters $\gamma = f_p$ and the $\xi = d / f_s$, we have (\ref{eqn:ucm}), the UCM. Additionally, if we constrain the pinhole camera plane to be on the surface of the sphere (\ie, $d = f_s$), and make $f_p = 2f_s$, we get the stereographic equation (\ref{eqn:stereographicradial}).\par

The E-UCM \cite{khomutenko2016eucm} extended the UCM by projecting to an ellipsoid instead of a sphere. This type of projection is known in geodesy and cartography for a long time \cite{laubscher1965map, snyder1987maps} as \textit{ellipsoidal} general perspective projections. We will not re-derive the equations here but would refer the reader to the source material. As mentioned, the DS model \cite{usenko2018doublesphere} extends the UCM by adding a second projection sphere to model more complex optics.\par

Thus, the UCM, the E-UCM and the DS models of fisheye lenses can be considered as generalizations of the stereographic camera model. It may be even more correct to say that they all (UCM, E-UCM, DS, division, and stereographic models) are part of a class of general perspective models. If we allow $f_s$ to approach infinity, then (\ref{eqn:generalperspective}) becomes the pinhole projection model. If we allow $f_p$ (and thus also $d$) to go to infinity, then we get the orthographic projection. \par
% -------------------------------------------------
\begin{figure}
  \captionsetup{font=footnotesize, belowskip=0pt}
  \centering
  \includegraphics[width=\linewidth]{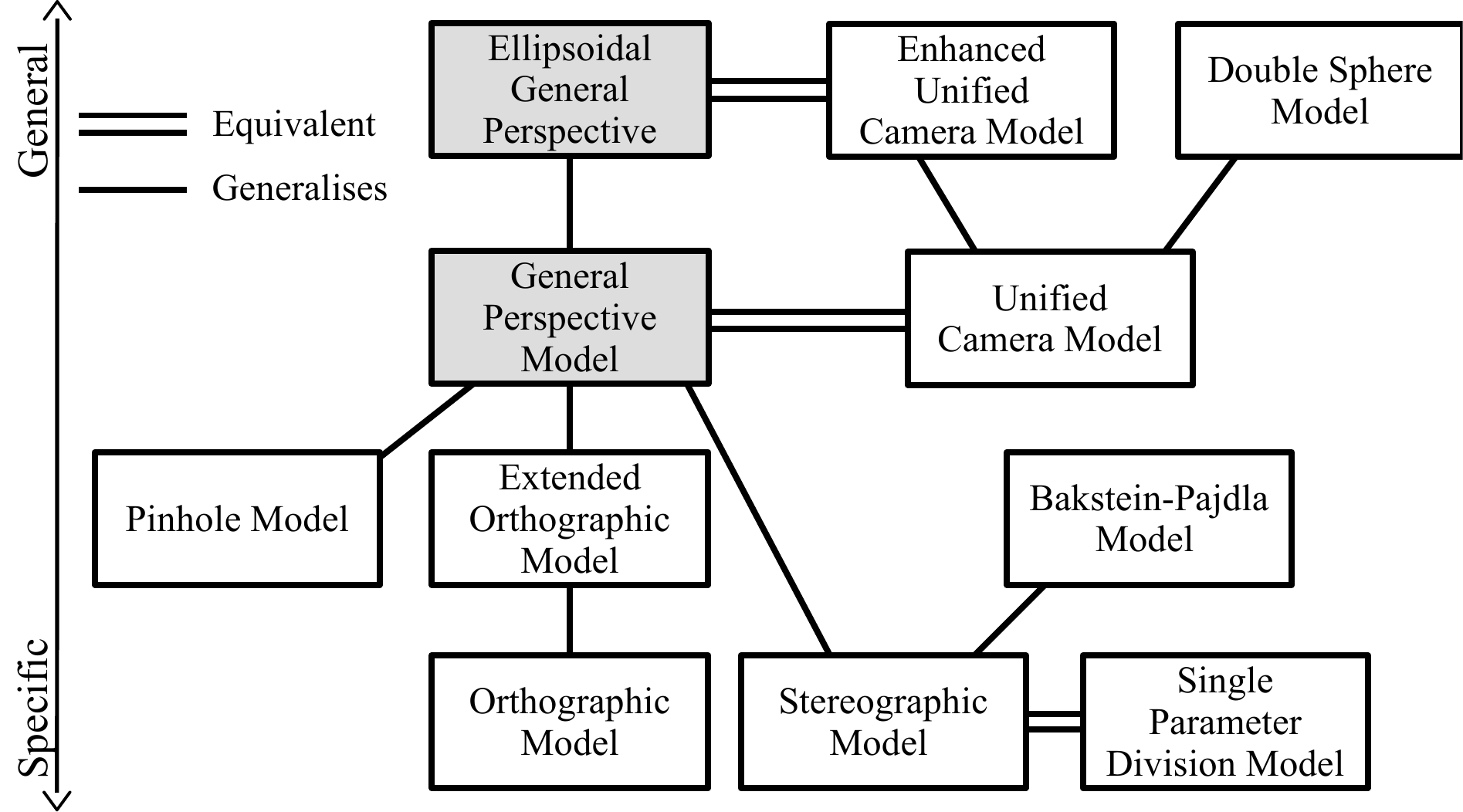}
  \vspace{-0.5cm}
  \caption{\textbf{The relationship between the various fisheye models and the general perspective projections.} Double line indicates that two models are equivalent, and single line indicates a generalisation/specialisation. }
  \label{fig:relationships}
  \vspace{-0.4cm}
\end{figure}
% -------------------------------------------------
\textbf{Stereographic and Division Models:}
As discussed in \cite{hughes2010fisheyeaccuracy}, We can combine the pinhole projection (\ref{eqn:pinholeradial}) with the inverse of the stereographic model (\ref{eqn:stereographicradialinv}) to give
% -------------------------------------------------
\begin{equation}
\onimg_{s}(\|\vec{u}_d\|) = f\tan{\left(2\atan{\left(\frac{\|\vec{u}_d\|}{2f}\right)}\right)} = \frac{\|\vec{u}_d\|}{1 - \frac{\|\vec{u}_d\|^2}{4f^2}}
\end{equation}
% -------------------------------------------------
Allow $a = 1 / 4f^2$, this is the same as the division model, (\ref{eqn:divisionmodelradialund}). Thus, we can say that the division model is the on-image version of the stereographic projection.\par
% -------------------------------------------------
\textbf{Equidistant and Field-of-View Models:}
Consider the radial pinhole projection given by (\ref{eqn:pinholeradial}), and the equidistant fisheye projection model (\ref{eqn:equidistantradial}). Combining the two to a similar form as the field-of-view model (\ref{eqn:fovmodelradial})
% -------------------------------------------------
\begin{equation} \label{eqn:equiundist}
\onimg_{e}(\|\vec{u}_d\|) = f_p\tan\frac{\|\vec{u}_d\|}{f_e}
\end{equation}
% -------------------------------------------------
As $f_p$ and $f_e$ are free parameters, determined through calibration, we can set them to
\begin{equation} \label{eqn:fov2equi}
f_p = \frac{1}{2\tan\frac{\omega}{2}} \;\;\text{and}\;\; f_e = \frac{1}{\omega}
\end{equation}
% -------------------------------------------------
Thus we see that (\ref{eqn:fovmodelradial}) and (\ref{eqn:equiundist}) are equivalent mapping functions. The field-of-view model is the on-image version of the equidistant projection.\par
% -------------------------------------------------
\textbf{Results:} To concretely demonstrate the equivalence of the Stereographic/Division and Equidistant/Field-Of-View model pairs, we provide a small set of results. Usefully, a set of parameters for five cameras is provided in \cite{usenko2018doublesphere}, including parameters for the field-of-view model. %
Given the set of parameters $\omega$ for the field-of-view model from \cite{usenko2018doublesphere},
we obtain the equidistant parameters through applying (\ref{eqn:fov2equi}). See Table \ref{tab:equiParams}. 
The difference between the two is at the level of machine precision, demonstrating the equivalence of the two models.
% -------------------------------------------------
\begin{table}[t]
\captionsetup{font=footnotesize, belowskip=0pt}
\caption{\bf Field-of-view parameters and equivalent equidistant parameters.}
% \vspace{-0.5cm}
\begin{center}
\begin{tabular}{c c c c c}
\toprule
      & Cam 1 & Cam 2 & Cam 3 & Cam 4 \\
\midrule
$w$   & 0.93  & 0.92  & 0.95  & 0.90  \\
$f_p$ & 0.997 & 1.009 & 0.972 & 1.035 \\
$f_e$ & 1.075 & 1.087 & 1.053 & 1.111 \\
Max error ($\times 10^{-14}$) & $0.4$ & $0.2$ & $0.4$ & $0.2$ \\
\bottomrule
\end{tabular}
\label{tab:equiParams}
\end{center}
\vspace{-0.4cm}
\end{table}
% -------------------------------------------------
The equivalence of the stereographic and division models is supported by the results presented in \cite{Courbon2012FisheyeAnalysis} (in particular, reference Table 4). We can see there that there is zero residual when the stereographic model is compared to the division model. We can also see that there is zero residual when the UCM (called USM in \cite{Courbon2012FisheyeAnalysis}), or equivalently the General Perspective Mapping, is compared to the stereographic and the orthographic models.
% -------------------------------------------------

\textbf{Discussion:}
There is a great number of potential models for application with fisheye cameras. In this paper, we have mentioned \textit{twenty} models, though for sure this is not exhaustive.
However, we have shown that there is a strong relationship among many of the geometric models. At least seven of the models are related to or directly equivalent to the General Perspective Projection. In addition, we have shown that some of the more recently developed fisheye models are mathematically equivalent to the classical fisheye projection functions, being the stereographic and the equidistant models proposed decades ago. In Figure \ref{fig:relationships}, we provide a map of geometric fisheye models that are related to the General Perspective Projection. \par
% -------------------------------------------------

% SVS system
% -------------------------------------------------
\section{Surround View Camera System} 
\label{sec:svs}

In this section, we discuss the setup of Surround View Cameras (SVC) and its basic primitives which are necessary for perception. We start with the historical usage of SVC for visualization which provides an understanding of the automotive configuration. We then discuss the supporting modules such as calibration, rectification, and geometric primitives.\\
% -------------------------------------------------
\begin{figure}[t]
 \captionsetup{font=footnotesize, belowskip=0pt}
  \centering
\begin{subfigure}{0.45\textwidth}
  \centering
  \includegraphics[width=\linewidth]{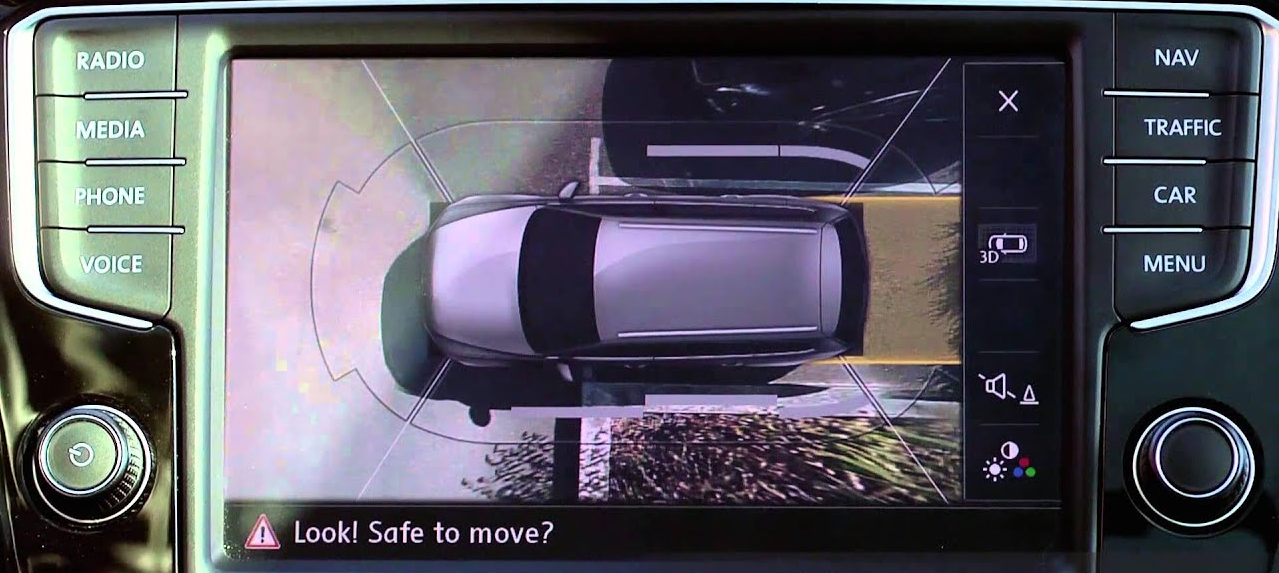}
 \subcaption{2D top view rendering.}
\end{subfigure}

\vspace{1mm} % don't delete white space above and below, it causes issues

\begin{subfigure}{0.45\textwidth}
  \centering
  \includegraphics[width=\linewidth]{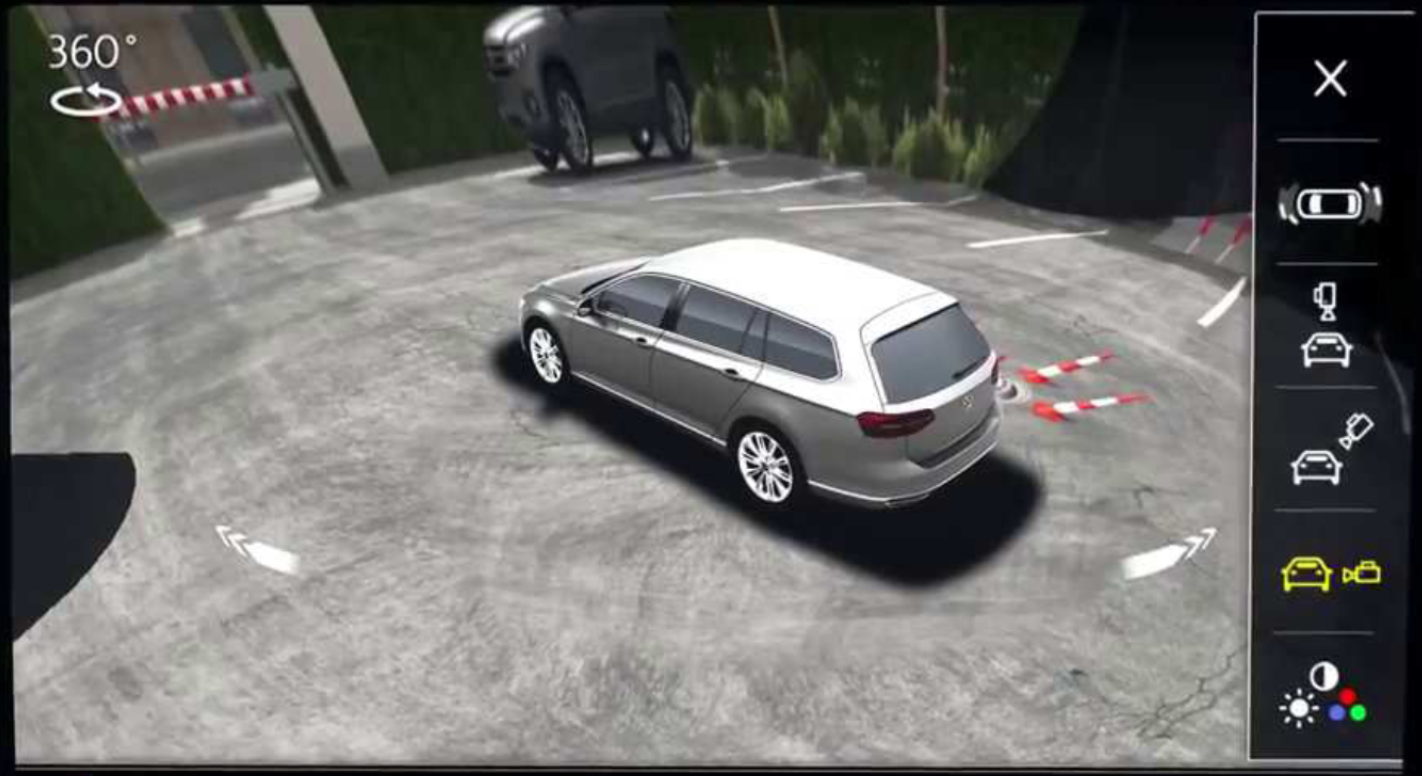}
  \subcaption{3D bowl view rendering.}
\end{subfigure} 
  \caption{\bf Surround-View camera visualization. }
  \label{fig:3D-surround-view}
\vspace{-0.4cm}
\end{figure}
% -------------------------------------------------
\begin{figure*}[ht]
    \captionsetup{font=footnotesize, belowskip=0pt}
    \centering
    \includegraphics[width=1.6\columnwidth]{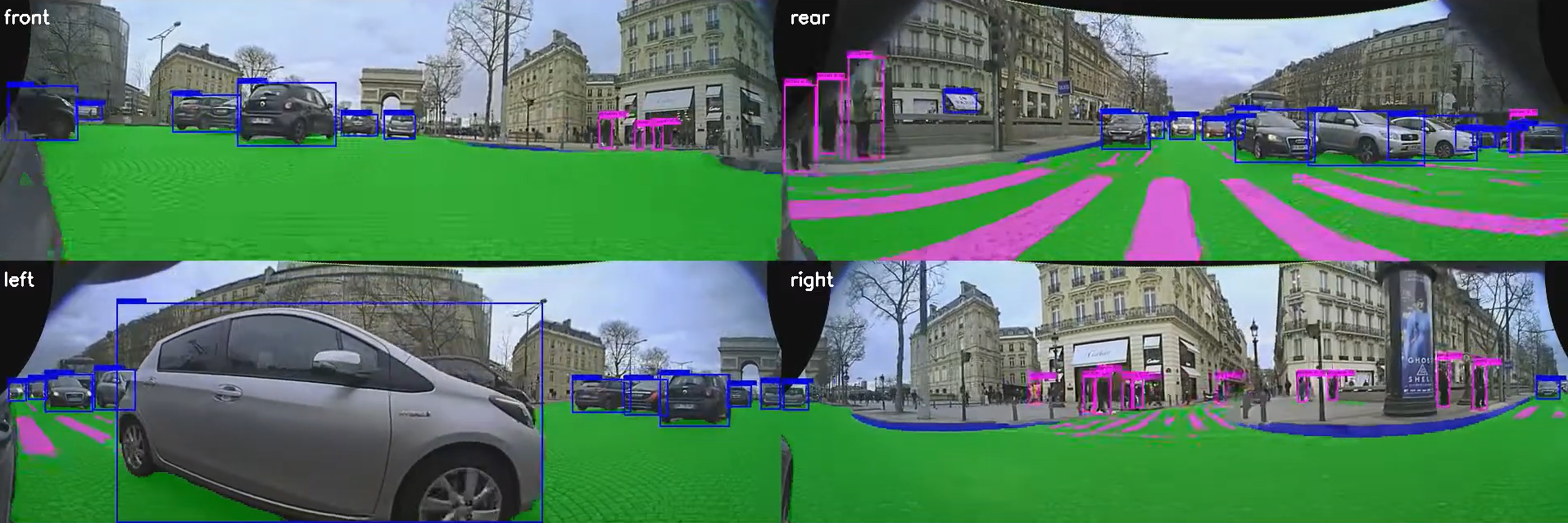}
    \caption{\textbf{Illustration of object detection and segmentation on cylindrical rectified surround-view images from a commercially deployed system~\cite{eising2021near} (See \url{https://youtu.be/ae8bCOF77uY} for a full video).} Overlapping field-of-view can be observed by noticing the arched gate and the vehicle in front of it. }\label{fig:od-fov}
\vspace{-0.4cm}
\end{figure*}
% -------------------------------------------------
\textbf{Visualization:} \label{subsec:svs-vis}
SVC have been historically used for display on the driver's dashboard for parking visualization. The first visualization application displayed a rear-view fisheye camera for reversing assist \cite{hughes2009wide}. It was further improved by visualization of object detection like pedestrians and the driving tube path \cite{Calefato2017} and was subsequently enhanced into surround-view visualization using four fisheye cameras \cite{Dabral2014}. Initial systems were based on a 2D top view as shown in Figure \ref{fig:3D-surround-view} (a). This was mainly used for parking applications, but it could also be used for other low-speed maneuvering use cases like traffic jam assist. 2D top view assumed a flat ground, and thus it had artefacts when the ground surface had a non-flat profile. Other nearby objects, such as vehicles, were heavily distorted in this view. They were resolved by a 3D surround view that uses a bowl-like surface that is flat nearby and has an upward curvature towards the periphery, as shown in Figure \ref{fig:3D-surround-view} (b). In addition, depth estimation around the vehicle can be used to adapt the bowl shape for optimal viewing with lesser artifacts of nearby objects. For example, if a vehicle is nearby on one side, the bowl surface in that region is brought in front of the vehicle to avoid artifacts. Typically, the application provides a user interface to select a viewpoint needed by the driver dynamically. Surround-view visualization application is usually implemented as an OpenGL \cite{woo1999opengl} rendering application using a graphics processing unit (GPU).\par
% -------------------------------------------------
Classically, imaging pipelines for SVC systems are designed primarily for visualization. However, this is sub-optimal for computer vision and a dual image pipeline was proposed in \cite{yahiaoui2019optimization, yahiaoui2019overview}. The control loop part of the image pipeline such as auto-exposure and auto-gain control is typically tuned for visualization as they cannot be jointly tuned. As the four SVC point in four different directions, they may have different ambient lighting. For example, sun rays may be hitting on the front of the vehicle and the corresponding image has high sun glare and saturation. Whereas the rear camera has the corresponding shadows and is dark. To improve the visual quality, the image brightness and color are harmonized when they are stitched together \cite{zlokolica2017free}. This could affect computer vision if the harmonization is done in the common image pipeline for visualization and computer vision.\par
% -------------------------------------------------
\textbf{Configuration:}  \label{subsec:config}
The main motivation for using fisheye cameras in an SVC system is to cover the entire $360^\circ$ near-field region around the vehicle. This is achieved by four fisheye cameras with a large horizontal field-of-view (hFOV) of around $190^\circ$ and a vertical field-of-view (vFOV) of around $150^\circ$. A fisheye camera has a very large angular volume coverage, but its angular resolution is relatively small, and it cannot perceive smaller objects at long range. Thus, it is primarily used as a near-field sensor. For comparison, a typical far-field front camera has hFOV of $120^\circ$ and vFOV of $60^\circ$. The angular volume is significantly smaller, but it has a much higher angular resolution enabling it to perceive objects in far range. The large hFOV of fisheye cameras enables $360^\circ$ coverage with only four fisheye cameras. The large vertical field-of-view enables capturing the region close to the vehicle, \eg, detection of higher elevation objects like a traffic light when stopped at a junction.\par
% -------------------------------------------------
Figure \ref{fig:surroundview} shows the mounting positions and the field-of-views of a typical SVC system. Four cameras are placed on four sides of the car marked with a blue circle for their positions. The front camera is placed on the front grille of the car and the rear camera is typically on the boot door handle. Left and right-side cameras are placed under the wing mirrors. Together they cover the entire $360^\circ$ region around the vehicle. The cameras are placed in such a way that the region very close to the vehicle is visible, which is crucial for parking scenarios. Because of this, a significant portion of the camera includes the ego vehicle. One can also notice the significant overlap of the field-of-view as seen in the intersecting regions. This can be exploited to resolve scale in structure from motion problems. However, this overlap is at the periphery which has the highest distortion, and it is hard to get algorithms to work accurately in this region.
Figure \ref{fig:od-fov} illustrates object detection and segmentation on a commercially deployed near-field perception system \cite{eising2021near} tested on a busy urban street in Paris. The overlapping field-of-view can be observed by noticing the arched gate which is seen in the center of the front camera and on the edges of the left and right cameras. The silver car in front of the gate is detected in all three cameras. Very wide-angle detections of vehicles (left end of left camera) and pedestrians (left end of rear camera) can also be observed.\par
% -------------------------------------------------
\textbf{Calibration:}
Previously, we have discussed various models for fisheye cameras. Each of these models has a set of parameters (known as \textit{intrinsic parameters}, that must be estimated through a calibration procedure.
In addition, the \textit{extrinsic parameters} of the camera should be estimated, being the position and orientation of the camera system in the vehicle coordinate system
\cite{zhang1999,heikkila97}.
A typical calibration process is that, first, image features are being detected (\eg, corners in the check board pattern \cite{Duda2018AccurateDA}) and secondly an algorithm will try to estimate the intrinsic and extrinsic parameters to project the detected features using the model of the calibration setup, by minimizing the reprojection error of the points. The reprojection error indicates hereby how well a model having a set of parameters can represent the projection function of the lens. Other photogrammetric approaches use vanishing Point Extraction and sets lines for estimating the calibration parameters \cite{hughes2010equidistant,antunes2017unsupervised}.
% -------------------------------------------------
\begin{figure}[tb] 
    \captionsetup{font=footnotesize, belowskip=-8pt}
    \centering
    \includegraphics[width=\columnwidth]{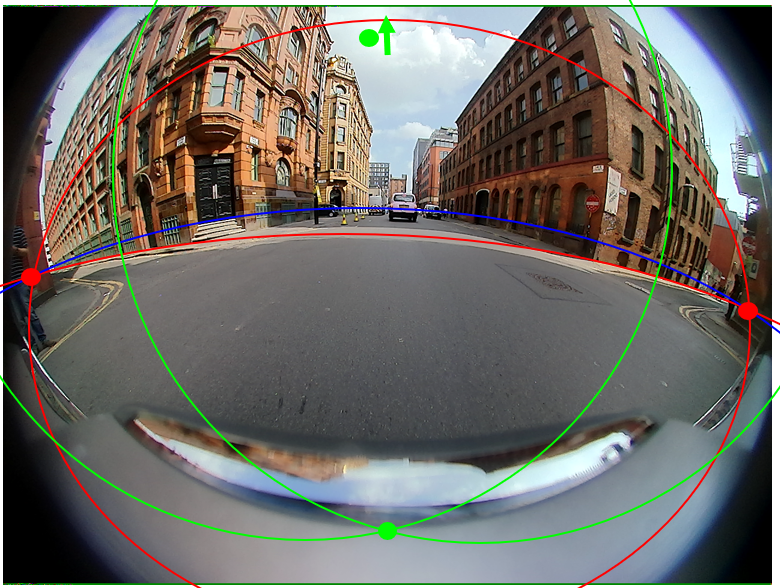}
    \caption{\textbf{Illustration of horizon line,  vanishing points, and epipolar lines.} Lines in the fisheye image can be approximated as conics. Equivalent to how parallel straight lines in a perspective image converge at a single vanishing point, parallel straight lines in a fisheye image converge at two vanishing points. The two vanishing points, when raised to the unit sphere, are antipodal points on the sphere. \textcolor{red}{Red} and \textcolor{darkgreen}{green} denote the perspective of horizontal parallel lines (with \textcolor{blue}{blue} as the associated horizon line) and vertical parallel lines respectively. The red and green dots denote the vanishing points with one of the vanishing points outside the image.}
    \label{fig:fe_persp}
\vspace{-2mm}
\end{figure}
% -------------------------------------------------
A widely known toolbox for calibration is implemented in the OpenCV library\cite{opencv_library}.
OpenCV also offers a version for fisheye camera models \cite{kannala2006fisheye} (\S\ref{sec:polynomialmodel}).
Other known calibration toolboxes for fisheye cameras are OCamCalib \cite{scaramuzza2006Toolbox, Scaramuzza2006Calibration, rufli2008} and Kalibr \cite{rehder2016, furgale2013, furgale2012, maye2013}. Finally, in \cite{Heng2013Calib}, a calibration process for multiple fisheye cameras on a vehicle is proposed (intrinsic and extrinsic), based on the extraction of checkerboard features and inter-camera correspondences. This is suitable for surround view camera systems, as it provides an accurate extrinsic calibration with respective to the vehicle is prerequisite for providing a seamless surround-view image. When the calibration patterns have a known position with respective to the vehicles coordinate system, the pose of the camera can be estimation like described above in off-line environment \cite{Shao2019}.
% -------------------------------------------------
Over the lifetime of a vehicle, the camera's pose relative to the vehicle can drift, due to wear of mechanical parts. It is desirable for the camera system to update its calibration automatically, with a class of algorithms. To correct the camera poses' change in online environments, it is possible to minimize photometric errors between ground projections of adjacent cameras \cite{Liu2019calib}. The approach of Choi \etal exploit corresponding lane markings captured and detected by adjacent cameras to refine an initial calibration \cite{choi2018calib}. 
In \cite{Zhanpeng2020calib} Ouyang \etal, a strategy to optimize the exterior orientations by taking the vehicle odometry into account is presented, by estimating the vehicle forward movement using geometry consistency and the vehicle direction using the vertical vanishing point estimates. Those algorithms are mostly used to correct geometric misalignment, but require an initial location obtained by an offline calibration. Friel \etal \cite{friel2010calib} describe a method of automatically extracting fisheye intrinsics from an automotive video sequence using, though it is limited to single parameter fisheye models (such as the equidistant model). 
% \sy{Lenses of the surround view cameras are directly exposed to external environment unlike front camera which is under the windshield. External environmental temperature changes can affect the focal length of the camera and thus it must be handled in online calibration for higher accuracy.}
% -------------------------------------------------

\textbf{Projection geometry:}
In a pinhole camera, any set of parallel lines on a plane converge at a single {\em{vanishing point}}. Those can be used to estimate the intrinsic and extrinsic parameters. For pinhole camera models, geometric problems can often be formulated using linear algebra. In this case parallel lines can be detected using a Hough-Transformation \cite{Aggarwal2006}. The set of all vanishing points is the {\em{horizon line}} for that plane. In a real world camera system, the pinhole camera is a mathematical model of the camera, that has errors in the form of, \eg, optical distortions. This is generally acceptable for narrow field-of-view cameras, where the distortion is mild. For wide field-of-view cameras, the distortion is too great for this to be a practical solution, and if the field-of-view (FOV) of the camera is greater than $180^\circ$, then there is not a one-to-one relationship of points in the original image to the corrected image plane. For fisheye cameras, a better model is the spherical projection surface \cite{Forstner2016book, mariotti2020spherical}. In the fisheye image, Hughes \etal describe in \cite{hughes2010equidistant}, how those parallel lines can be approximated and fitted as circles or conics for fisheye cameras to determine vanishing points or horizontal lines. These parallel lines correspond to great circles of the spherical surface. Correspondingly, straight lines imaged by a fisheye camera are approximately conic \cite{hughes2010fisheye}, and parallel lines imaged by a fisheye camera converge at two vanishing points (Figure \ref{fig:fe_persp}). \par
\begin{figure}[!t]
  \captionsetup{font=footnotesize, belowskip=0pt}
  \centering
  \includegraphics[width=\linewidth]{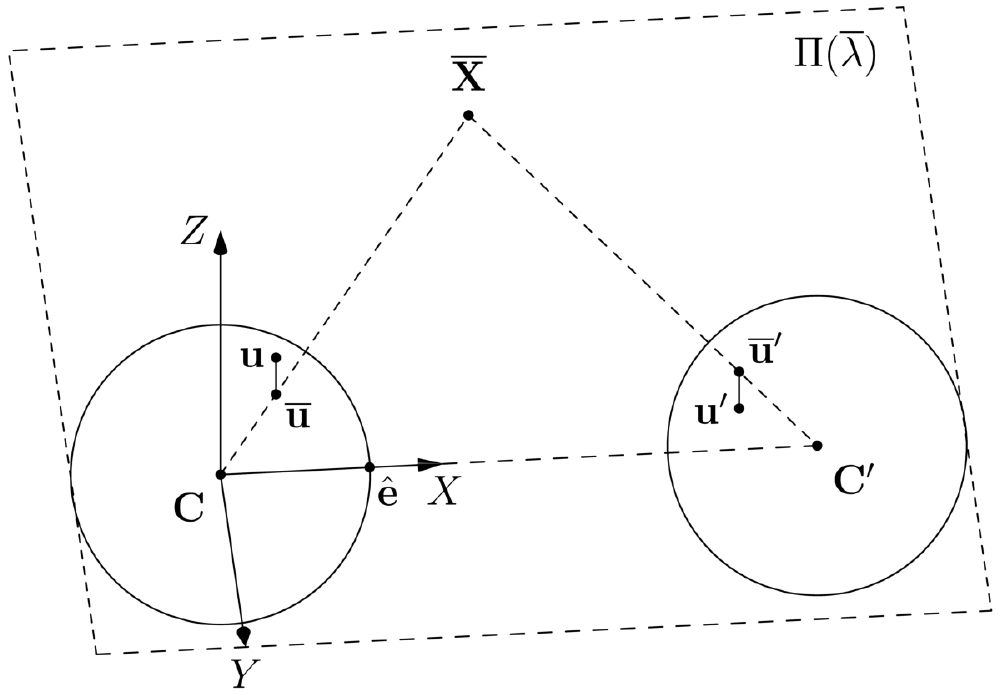}
  \caption{\textbf{Spherical epipolar geometry.} The epipolar plane $\Pi$ is one of the pencil of planes about the epipole $\vec{e}$, defined by the camera centers $C$ and $C'$. Ideal observations $\overline{\vec{u}}$ and $\overline{\vec{u}}'$ will lie on the epipolar plane. However, actual observed points $\vec{u}$ and $\vec{u}'$, in the presence of noise, will have a non-zero distance to the epipolar plane.}
  \label{fig:sphrepi}
\end{figure}

% -------------------------------------------------
\textbf{Spherical Epipolar geometry:}
The geometric relations of stereo vision are described by {\em{epipolar geometry}}, which can be used for {\em{depth estimation}} and {\em{structure from motion}} approaches in combination with feature extractors.
In pinhole camera models, the intersection of the line passing through the two camera optical centers and the image planes define special points called {\em{epipoles}}. This line is called the {\em{baseline}}. Each plane through the baseline defines matching epipolar lines in the two image planes. A point in one camera is located on an epipolar line on the other and vice versa. This reduces the search of a corresponding point (stereo matching) in a two-view camera setup to a 1D problem. For omnidirectional cameras, such as fisheye, where we employ spherical projection surfaces in place of planar, it is more intuitive to discuss epipolar planes instead of epipolar lines, as described in Figure \ref{fig:sphrepi}. Ideal observations of a single 3D point from two cameras will lie on the same epipolar plane, in the same way that they lie on the epipolar lines in the pinhole case. It is important, however, to note that the cameras must be calibrated to raise image features to the projective sphere. In contrast, for narrow FOV cameras, epipolar geometry is defined for the uncalibrated case, via the fundamental matrix.\par

% -------------------------------------------------
% -------------------------------------------------

% -------------------------------------------------
\begin{figure}[tb]
  \captionsetup{font=footnotesize, belowskip=0pt}
    \centering
    \includegraphics[width=\columnwidth]{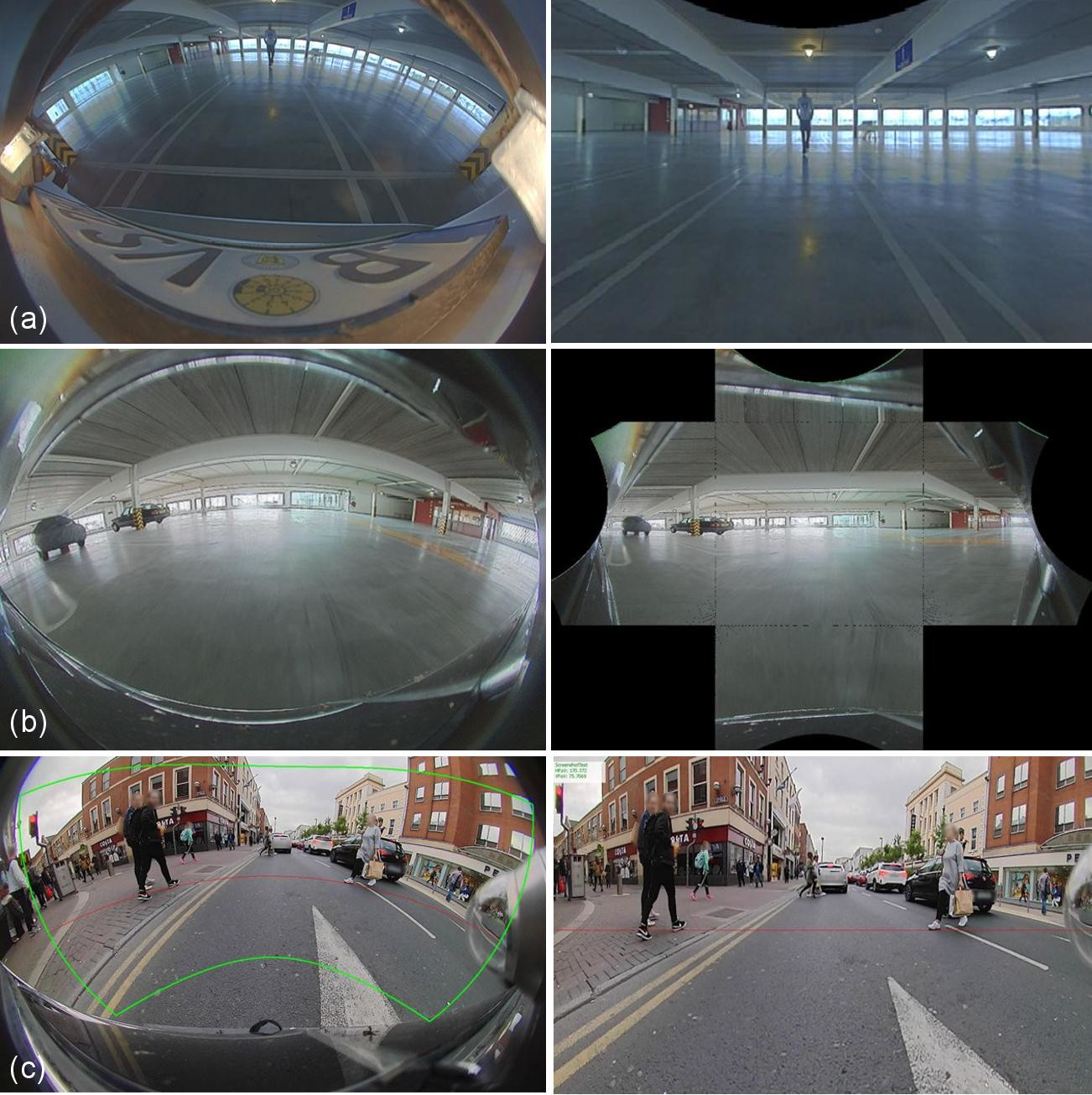}
    \caption{\textbf{Undistortion of fisheye images:} 
    (a) Rectilinear correction; (b) Piecewise linear correction; (c) Cylindrical correction. Left: raw image; Right: undistorted image.}
    \label{fig:rectify}
\vspace{-0.4cm}
\end{figure}
% -------------------------------------------------
\textbf{Rectification:} \label{sec:rectification}
It is possible to remove radial distortion in fisheye cameras and re-use standard perception algorithms. Although it is a rapid way to start fisheye camera perception development, there are several problems associated with rectification. Firstly, it is theoretically impossible to rectify a fisheye image to a rectilinear viewport as the horizontal field-of-view is greater than $180^\circ$, thus there are rays that are incident on the lens behind the camera which doesn't work for pinhole setup. It is counterproductive to use a fisheye lens with a large field-of-view and then lose some of it because of rectification.
The second significant problem is resampling distortion which is more practical in nature. It is a particular manifestation of interpolation artifacts, wherein for fisheye images, a small patch (particularly at the periphery where the distortion is high) is expanded to a very large region in the rectified image leading to high noise. In particular, the negative impact on computer vision due to the introduction of spurious frequency components by resampling is discussed in~\cite{LourencoSIFT}. Additionally, the warping step is needed at inference time, which consumes significant computing power and memory bandwidth. It creates a non-rectangular image with invalid pixels which reduces computational efficiency further.\par

Commonly used rectification methods for fisheye are shown in Figure \ref{fig:rectify}. Figure \ref{fig:rectify} (a) shows the standard rectilinear correction. Significant loss of near field can be observed from the missing horizontal white line. Regions at the left and right edges are missing as well. Although there is a significant loss, this enables the usage of standard camera algorithms. Figure \ref{fig:rectify} (b) shows a cubic approximation where the fisheye lens manifold surface is approximate by an open cube. It can be interpreted as a piecewise linear approximation of the fisheye projection surface. Each plane is a rectilinear correction and hence standard algorithms can be used within each block. However, the distortion across two surfaces of the cube has a large distortion and it will be difficult to detect objects which are split across the two regions. One can also notice the strong perspective distortion and blurriness due to re-sampling artifacts at the periphery.\par
% -------------------------------------------------
Practically, a common rectification process is to use a cylindrical surface as illustrated in Figure \ref{fig:rectify} (c). It can be interpreted as a quasi-linear approximation as it is linear in the vertical direction and the surface has a quadratic curvature in the horizontal direction. It covers a significantly larger field-of-view relative to a rectilinear viewport. The main advantage is that the vertical objects remain vertical as observed by vertical lines on the building \cite{Plaut_2021_CVPR}. Thus, scanlines are preserved for performing searches horizontally for stereo algorithms between two consecutive fisheye images (motion stereo) or between a fisheye and a narrow field-of-view camera (asymmetric stereo). The main disadvantage is its inherent inability to capture the near field region close to the vehicle. This can be fixed by using an additional smooth surface covering the near-field region. There is also an increased distortion of nearby objects. \par
% Perception Tasks
% -------------------------------------------------
\section{Perception Tasks} 
\label{sec:tasks}

There is relatively less literature on perception tasks for fisheye images as there are limited datasets. We split the perception tasks into semantic, geometric, and temporal tasks. Finally, we discuss joint multi-task models.

\subsection{Semantic Tasks}

In this section, we discuss semantic tasks which are based on appearance-based pattern recognition.\par
% -------------------------------------------------
\begin{figure*}[!t]
  \captionsetup{font=footnotesize, belowskip=0pt}
    \centering
    \includegraphics[width=0.8\textwidth]{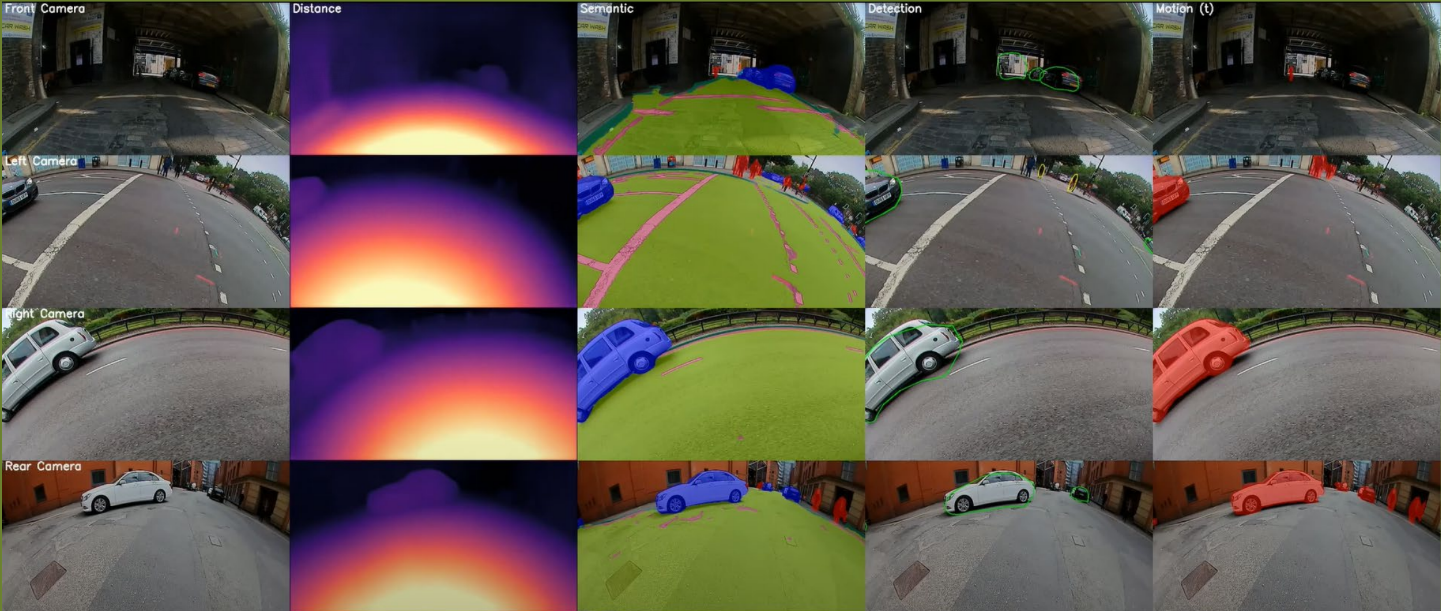}
    \caption{{\bf Qualitative results of raw fisheye images from the OmniDet framework on the WoodScape dataset \cite{yogamani2019woodscape}.}
    The 1\textsuperscript{st} column indicates the input images from Front, Left, Right and Rear cameras, 2\textsuperscript{nd} column indicate distance estimates, 3\textsuperscript{rd} column semantic segmentation maps, 4\textsuperscript{th} indicate generalized object detection representations and 5\textsuperscript{th} indicate the motion segmentation. For more qualitative results at a higher resolution, check this video: \url{https://youtu.be/xbSjZ5OfPes}.}
    \label{fig:qualitative-results}
\vspace{-0.4cm}
\end{figure*}
% -------------------------------------------------
\textbf{Semantic Segmentation:}
It is the process of assigning a class label to each pixel in an image such as a pedestrian, road, or curb as shown in Figure~\ref{fig:qualitative-results} (2\textsuperscript{nd} column). CNN-based approaches have recently been very successful compared to classical computer vision-based methods on semantic segmentation employed on a pinhole front camera \cite{segdasvisapp19}. Although, in urban traffic scenarios, autonomous cars require a wider field-of-view to perceive what is around them, particularly at intersections. An Overlapping Pyramid Pooling module (OPP-Net) was presented by Deng~\etal~\cite{deng2017cnn} by employing multiple focal lengths to generate various fisheye images with their respective annotations. The OPP-Net was trained and evaluated on an existing urban traffic scene semantic segmentation dataset on fisheye images. Furthermore, to improve the model's generalization performance \cite{deng2017cnn} proposed a novel zoom augmentation technique to augment the data specifically designed for fisheye images. Extensive experiments indicated the effectiveness of the zoom augmentation technique and the OPP-net performed well in urban traffic scenarios. Saez~\etal~\cite{saez2019real} introduced a real-time semantic segmentation technique which is an adaptation of Efficient Residual Factorized Network (ERFNet)~\cite{romera2017erfnet} to fisheye road sequences and generated a new semantic segmentation dataset for fisheye cameras based on CityScapes \cite{cordts2016cityscapes}. The tests were performed on authentic fisheye sequences, although only qualitative results were revealed as there is no ground truth.\par

Deng~\etal~\cite{deng2019restricted} uses surround-view cameras to tackle 360$\degree$ road scene segmentation as they are widely adopted in production vehicles. To deal with the distortion issues in fisheye images, Restricted Deformable Convolution (RDC) was proposed. They allow effective geometric transformation modeling by learning the shape of the convolutional filter based on the input feature map. Also, the authors presented a zoom augmentation technique for converting perspective images into fisheye images. This facilitates in the creation of a large-scale training set of surround-view camera images. An RDC-based semantic segmentation model is also developed. A multi-task learning (MTL) architecture is used to train for real-world surround-view camera images by combining real-world and transformed images. These models were trained on Cityscapes~\cite{cordts2016cityscapes}, FisheyeCityScapes~\cite{ye2020universal} and SYNTHIA~\cite{ros2016synthia} datasets and tested on authentic fisheye images.\par

Clément~\etal~\cite{playout2021adaptable} focuses on showing that deformable convolutions can be used on top of an existing CNN without varying its pre-trained weights. This helps systems that bank on multiple image modalities as each model can now be modified reliably without retraining them from scratch. They also demonstrate that the deformable components can be trained independently (although if finetuning, it is recommended to use batch normalization). Authors also say that the need for large datasets of labeled fisheye images is eliminated. After learning the deformable components, fine-tuning network weights are not necessary for achieving high performance.\par

Yaozu~\etal~\cite{ye2020universal} presented a 7-degrees-of-freedom (DoF) augmentation technique for converting rectilinear perspective images into fisheye images. It includes a spatial relationship between the world and the fisheye coordinate system (6-DoF), as well as the virtual fisheye camera's focal length variations (1-DoF). During the training phase, rectilinear perspective images are turned into fisheye images in 7-DoF to replicate fisheye images taken by cameras with various locations, orientations, and focal lengths. This improves the model's accuracy and robustness while dealing with distorted fisheye data. The 7-DoF augmentation provides a generic solution for semantic segmentation for fisheye cameras as well as provides definite parameter settings for augmentation of autonomous driving and created the FisheyeCityScapes~\cite{ye2020universal} dataset.\par
% -------------------------------------------------
\textbf{Object Detection:}
Object detection suffers the most from radial distortion in fisheye images. Due to inherent distortion in the fisheye image formation, objects at different angles from the optical axis appear very different, making object detection difficult (see Figure \ref{fig:qualitative-results}). The rectangular bounding boxes tend to be not the best representation of the size of the object sometimes is half the standard bounding box whereas the box itself would be twice the object of interest. Instance segmentation provides precise contours of the objects, but they are much more expensive to annotate and require a bounding box estimation step. Rectification provides a significant benefit but it also suffers from the side effects discussed in Section \ref{sec:svs}.\par

FisheyeDet~\cite{li2020fisheyedet} underlines the need for a useful dataset. They create a simulated fisheye dataset by applying distortions to the Pascal VOC dataset~\cite{everingham2010pascal}. 4-sided polygon representation along with distortion shape matching helps FisheyeDet. The No-prior Fisheye Representation Method (NPFRM) was proposed to extract adaptive distortion features without using lens patterns and calibration patterns. Also, the Distortion Shape Matching (DSM) strategy was put forward to localize objects tightly and robustly in fisheye images. They use improper quadrilateral bounding boxes formed from the contour of distorted objects. An end-to-end network detector is created by combining it with the NPFRM and DSM.\par

SphereNet~\cite{coors2018spherenet} and its variants~\cite{perraudin2019deepsphere, su2019kernel, jiang2019spherical} formulate CNNs on spherical surfaces and also explicitly encode invariances against the distortions. SphereNet accomplishes this by reversing distortions by adapting the sampling locations of the convolutional filters and wrapping them around the sphere. Existing perspective CNN models can be transferred to the omnidirectional scenario using SphereNet, which is modeled on normal convolutions. Moreover, quasi distortion in both horizontal and vertical directions indicates that fisheye images do not conform to spherical projection models. The outcomes of several detection algorithms that utilize equirectangular projection (ERP) sequences as direct input data were compared by Yang~\etal~\cite{yang2018object}, revealing that the CNN produces only a certain accuracy without projecting ERP sequences into normal 2D images.\par

FisheyeYOLO \cite{rashed2021generalized, rashedfisheyeyolo} investigates various representations such as orientated bounding box, ellipse, and generic polygon. Using the intersection-over-union (IoU) metric and accurate instance segmentation ground truth, they compare these representations. They suggest a new curved bounding box approach with the best features for fisheye distortion camera models, as well as a curvature adaptive perimeter sampling methodology for generating polygon vertices that enhances relative mAP score by 4.9\% over uniform sampling. Overall, the suggested polygon CNN model improves mean IoU relative accuracy by 40.3\%.\par
% -------------------------------------------------
\textbf{Soiling:} Surround view cameras are directly exposed to the external environment and is susceptible to soiling. For comparison, a front camera is placed behind the windshield, and it is less susceptible. This task was first formally defined in ~\cite{uvrivcavr2019soilingnet}. There are two types of soiled areas: opaque (mud, dust, snow) and transparent (water, oil, and grease) (water). Transparent soiling might be difficult to identify due to the limited visibility of the background. Soiling can cause significant degradation of perception accuracy, thus cleaning systems using a spray of water or more advanced ultrasonic based cleaning is employed for higher level of automated driving. Even if the camera is not cleaned, soiling detection is required to enhance the uncertainty of vision algorithms in degraded areas. 
As it is difficult to collect soiled data, DirtyGAN \cite{uricar2021let} proposed to use generative adversarial networks (GANs) to artificially generate different soiling patterns inpainted on real scenes. Boundaries of soiling are blurry and not well defined; thus, the manual annotation can be subjective and noisy. Das \etal \cite{das2020tiledsoilingnet} proposed tile level soiling classification to handled noisy annotations and to improve computational efficiency. Uricar \etal \cite{uvrivcavr2021ensemble} proposed to use an ensemble based semi-supervised learning of pseudo labels to refine the noisy annotations automatically.\par

From a perception perspective, there are two ways to handle soiling. One way is to include robustness measures to improve the perception algorithm. For \eg, Sakaridis \etal\cite{sakaridis2018semantic} proposed a foggy scene aware semantic segmentation. The other way is to restore the soiled region. Mud or water droplets are generally static or occasionally have low-frequency dynamics of moving water droplets. Thus, it is more effective to use video-based restoration techniques. 
Porav \etal \cite{porav2019can} explored transparent soiling by using a stereo camera in conjunction with a dripping water source to mimic raindrops on the camera lens. This was done to automatically annotate the rainy pixels and they trained a CNN to restore the rainy regions. A desoiling dataset benchmark for surround view cameras was provided by Uricar \etal \cite{uricar2019desoiling}. They use three cameras in proximity with various levels of soiling and a fourth camera with no soiling which acts as ground truth. They implemented a multi-frame baseline which can profit from the visibility of soiling occluded zones as time passes.\par

Sun glare detection is a closely related task of soiling. In manual and automatic driving, glare from the sun is a typical issue. Overexposure in the image is caused by sun glare, which substantially impacts visual perception algorithms. It is critical for higher levels of autonomous driving for the system to recognize that there is sun glare, which can degrade the system. The literature on detecting sun glare for automated driving is scarce. It is essentially based on image processing algorithms to detect saturated brightness areas and extract sections. A highly resilient algorithm is required from the perspective of a safety system. As a result, Yahiaoui~\etal~\cite{yahiaoui2020let} created two complementary algorithms that use traditional image processing techniques and CNN to learn global context.\par
% -------------------------------------------------
\textbf{Chargepad Assist:}
Electric vehicles are becoming more widespread, and inductive chargepads are a practical and effective way to charge them. However, because drivers are usually poor at accurately aligning their vehicles for optimal inductive charging, a desirable solution would be to perform an automated alignment of the charging plates. The usage of surround-view cameras is ideal as it's a near-field perception task and implemented as an extension of the automated parking system. Dahal~\etal~\cite{dahal2021online} proposes a methodology premised on a surround-view camera framework that automatically identifies, localizes, and aligns the vehicle with the inductive chargepad. The visual design of the chargepads is not consistent and is often not recognized ahead of time. As a result, employing a system that depends on offline training would occasionally fail. 
Henceforth, they propose a self-supervised online learning technique that learns a classifier to auto-annotate the chargepad in the video sequences for further training by leveraging the driver's actions when manually aligning the car with the chargepad along with the weakly supervised semantic segmentation and depth predictions. When confronted with a previously undetected chargepad, the driver would have to align the car once manually as the chargepad lying on the ground is flat and is not easy to see and spot from afar. To achieve alignment from a more extensive range, they propose employing a Visual Simultaneous Localization and Mapping (SLAM) framework to learn landmarks relative to the chargepad.\par

\textbf{Trailer Assist:} 
Trailers are frequently used to move products and recreation equipment. Maneuvers with trailers, particularly reversing, can be tricky and unpleasant even for seasoned drivers. As a result, driver assistance systems come in handy in these situations. A single rear-view fisheye camera perception algorithm is usually used to achieve these. There is relatively little academic research on the subject because there is no publicly available dataset for this challenge. This prompted Dahal~\etal~\cite{dahal2019deeptrailerassist} to detail all the trailer assist use cases and suggest a CNN-based solution to trailer perception issues. Using deep learning, they built a dataset for trailer detection and articulation angle estimation tasks. They developed and obtained high accuracy by detecting and tracking the trailer and its angle with an efficient CNN and long short-term memory (LSTM) model.\par
% -------------------------------------------------
\subsection{Geometric Tasks}

\textbf{Depth Estimation:}
It involves estimating the distance to an object (or any plane) at a pixel level, as shown in Figure~\ref{fig:qualitative-results}. Calculating distance relative to a camera plane is still very challenging. Currently, most of the works are on the rectified KITTI~\cite{geiger2013vision} sequences where barrel distortion is removed. In the case of a pinhole camera, depth is defined as the perpendicular distance from the camera plane, namely $z$. Previous structure-from-motion (SfM) approaches \cite{zhou2017unsupervised, godard2019digging}, estimated inverse depth by parameterizing the network's disparity predictions into depth for the unprojection operation during the view synthesis step. This parameterization does not work well for fisheye cameras as they undergo large distortions which result in obtaining obtain angular disparities on the epipolar curves compared to the epipolar lines in the pinhole camera. To apply the same approach as pinhole, we would need to rectify the fisheye images which would result in a loss in field-of-view as described in Section \ref{sec:rectification}. However, the same multi-view geometry~\cite{hartley2003multiple} principles that apply to pinhole projection model cameras also apply to fisheye images. By observing the scene from differing viewpoints and establishing correspondences between them, the underlying geometrical structure can be estimated. It is noteworthy to consider the CNN to output norm values than angular disparities for fisheye cameras when \emph{SfM} approach is employed as it would make it difficult to parameterize the angular disparities to distance for the view synthesis operation. Furthermore, the value of $z$ can be (close to) zero or negative for field-of-views greater than $180^\degree$, which also leads to numerical issues as the models typically have some direct or indirect division by $z$ computation. Instead, it is useful to estimate the radial distance \ie norm $\sqrt{x^{2}+y^{2+}z^{2}}$ instead of $z$. The norm is always positive and non-zero (except for $x, y, z=0 $) and allows a more numerical stable implementation.\par

On LiDAR distance measurements, such as KITTI, depth prediction models can be learned in a supervised manner. Ravi Kumar~\etal~\cite{kumar2018near} took a similar method, demonstrating the ability to predict distance maps employing LiDAR ground truth for training on fisheye images. Although, LiDAR data is very sparse and expensive to set up with good calibration. To overcome this problem, FisheyeDistanceNet~\cite{kumar2020fisheyedistancenet} focused on solving one of the most challenging geometric problems, \ie, distance estimation on raw fisheye cameras using image-based reconstruction techniques, which is a challenging task, as the mapping between 2D images to 3D surfaces is an under-constrained problem. Depth estimation is also an ill-posed problem because there are several potential erroneous depths per pixel, which could also replicate the novel view. UnRectDepthNet~\cite{kumar2020unrectdepthnet} introduced a generic end-to-end self-supervised training framework for estimating monocular depth maps on raw distorted images for different camera models. The authors demonstrated the results of the framework work on raw KITTI and WoodScape datasets.\par

SynDistNet~\cite{kumar2020syndistnet} learned semantic-aware geometric representations that could disambiguate photometric ambiguities in a self-supervised learning \emph{SfM} context. They incorporated a generalized robust loss function~\cite{barron2019general}, which significantly improved performance while eliminating the necessity of hyperparameter tuning with the photometric loss. They employed a semantic masking approach to reduce the artifacts due to the dynamic objects that violated static world assumptions. SynDistNet considerably enhanced the root mean squared error (RMSE) when compared to prior methods~\cite{kumar2020fisheyedistancenet, kumar2020unrectdepthnet} on fisheye images, reducing it by 25\%. Most current depth estimation methodologies rely on a single camera, that cannot be seamlessly generalized to multiple fisheye cameras. Furthermore, the model must be implemented over several various-sized car lines with differing camera geometries. Even within a single-car line, intrinsics differ due to the manufacturing tolerances. Deep neural networks do seem to be sensitive to these changes, and training and testing each camera instance is nearly impossible. As a result, SVDistNet~\cite{kumar2021svdistnet} proposed an innovative camera-geometry adaptive multi-scale convolutions that use the camera parameters as a conditional input, allowing the network to generalize to previously unknown fisheye cameras.\par
% -------------------------------------------------
\textbf{Visual Odometry:}
Liu~\etal~\cite{LiuHSGP17} describes a conventional direct visual odometry technique for a fisheye stereo camera. The technique does both camera motion estimation and semi-dense reconstruction at the same time. There are two threads in the pipeline: one for tracking and one for mapping. They estimate the camera posture using semi-dense direct image alignment in the tracking thread. To circumvent the epipolar curve problem, the plane-sweeping stereo algorithm is used for stereo matching and to initialize the depth. Cui~\etal~\cite{cui2019real} demonstrated a large-scale, real-time dense geometric mapping technique using fisheye cameras. The camera poses were obtained from a global navigation satellite system/inertial navigation system (GNSS/INS) but they also propose that they can be retrieved from the visual-inertial odometry (VIO) framework. The depth map fusion uses the camera postures retrieved by these approaches. Heng~\etal~\cite{heng2016semi} described a semi-direct visual odometry algorithm for a fisheye stereo camera. In a tracking thread, they track-oriented patches while estimating camera poses; in a mapping thread, they estimate the coordinates and surface normal for every new patch to be tracked. Surface normal estimation allows us to track patches from distinct viewpoints. They do not employ descriptors or strong descriptors matching in their technique to detect patch correspondences. Instead, they employ photoconsistency-based approaches to find patch correspondences. Numerous visual odometry approaches for fisheye cameras, including \cite{geppert2019efficient} and~\cite{caruso2015large}, have recently been presented. In addition, Geppert~\etal~\cite{geppert2019efficient} used a multi-camera visual-inertial odometry framework to extend the visual-inertial localization technique for large-scale environments, resulting in a system that allows for accurate and drift-free pose estimation. Ravi Kumar~\etal~\cite{kumar2021omnidet} employed CNNs for the visual odometry task, which acts as an auxiliary task in the monocular distance estimation framework.\par
% -------------------------------------------------
\textbf{Motion Segmentation:} 
It is defined as the task of identifying the independently moving objects (pixels) such as vehicles and persons in a pair of sequences and separating them from the static background as shown in Figure~\ref{fig:qualitative-results}. It is used as an appearance agnostic way to detect arbitrary moving objects using motion cues that are not common like rare animals (\eg, kangaroo or a moose). It was first explored in MODNet~\cite{siam2018modnet} for autonomous driving. Recently, instance-level motion segmentation was defined and explored in InstanceMotSeg~\cite{mohamed2021monocular}. FisheyeMODNet~\cite{yahiaoui2019fisheyemodnet} extends it to fisheye cameras without rectification. There was no explicit motion compensation, but it was mentioned as future work. Mariotti~\etal~\cite{mariotti2020spherical} uses a classical approach to accomplishing this task based on vehicle odometry \cite{eising2021odom}. Spherical coordinate transformation of optical flow was performed and the positive height, depth, and epipolar constraints were adapted to work in this setup. They additionally propose anti-parallel constraint to remove motion parallax ambiguity which commonly occurs when a car is moving parallel to the ego-vehicle.\par
% -------------------------------------------------
\subsection{Temporal Tasks}

Although geometric tasks like depth and motion can use multiple frames for training and inference, the output is defined only on one frame. We define temporal tasks to be one whose output is defined on multiple frames. It typically requires multi-frame sequential annotation. 

\textbf{Tracking:} Object tracking is the common temporal task where an object has to be associated across multiple frames. Detection and tracking of moving objects were explored in~\cite{baek2018real} for surround-view cameras. They use a classical optical flow-based approach for tracking. WEPDTOF~\cite{tezcan2022wepdtof} is a recently released dataset for pedestrian detection and tracking on fisheye cameras in an overhead surveillance setup. Although it is not an automotive dataset, it captures the challenges necessary for developing a tracking system on fisheye cameras. Trajectory prediction is closely related to tracking where the location of the object of interest must be predicted for the next set of frames. In the case of autonomous driving, it is particularly done in 3D bird's eye view space. PLOP algorithm~\cite{buhet2021plop} explored doing trajectory prediction of vehicles on a fisheye front camera after applying cylindrical rectification.\par
% -------------------------------------------------
\textbf{Re-identification:}
Re-identification (Re-ID) is the association of detected objects across the cameras. It could also include association over time across cameras.
Wu~\etal~\cite{wu2020vehicle} propose to perform vehicle Re-ID on the surround view cameras and highlight the two significant challenges: Firstly, due to fisheye distortion, occlusion, truncation, and other factors, it is difficult to detect the same vehicle from previous image frames in a single-camera view. Secondly, in a multi-camera perspective, the appearance of the identical vehicle changes dramatically depending on which camera is used. They offer a new quality evaluation mechanism to counteract the effects of tracking box drift and target consistency. They employ a Re-ID network based on an attention mechanism, which is then paired with a spatial constraint method to improve the performance of diverse cameras.\par

Zhao~\etal~\cite{zhao2021pedestrian} proposes a pedestrian Re-ID algorithm. It consists of a single camera detection and tracking module and a two-camera ReID module applied to multi-camera views. It includes a single camera detection and tracking module as well as a two-camera ReID module for multi-camera views. Using a YOLOv3~\cite{YOLOV3}, the detection module recognizes pedestrians in single camera view videos. To track pedestrians and issue an ID to each identified pedestrian, the tracking model integrates OSnet~\cite{zhou2019omni} with DeepSORT~\cite{wojke2017simple}. Both models were adapted to the fisheye images using transfer learning procedures.\par
% -------------------------------------------------
\textbf{SLAM:}
Feature correspondence comprises keypoint detection, description, and matching and it is the primary step in SLAM systems. FisheyeSuperPoint~\cite{Konrad2021FisheyeSuperPointKD} introduces a unique training and evaluation methodology for fisheye images. As a starting point, they employ SuperPoint~\cite{detone2018superpoint}, a self-supervised keypoint detector and descriptor that has generated state-of-the-art homography prediction results. They present a fisheye adaption framework for training on undistorted fisheye images; fisheye warping is employed for self-supervised training on fisheye images. Through an intermediary projection phase to a unit sphere, the fisheye image is translated to a new, distorted image. The virtual posture of the camera can be changed in 6-Dof. Tripathi ~\etal \cite{tripathi2020trained} where they explored the problem of relocalization using surround-view fisheye cameras using an ORB SLAM pipeline. The goal was to perform mapping of a private area like an apartment complex and relocalize with respect to this map to assist automated parking. Feature detection was performed on raw fisheye images, and a comparison of different feature correspondence algorithms on raw fisheye cameras was analyzed.\par
% -------------------------------------------------
\begin{table*}[ht]
\captionsetup{font=footnotesize, belowskip=0pt}
\centering
\caption{\bf Summary of various autonomous driving datasets containing fisheye images.} 
\label{tab:datasets}
\begin{tabular}{@{}lcccccll@{}}
\toprule
Datasets & \textit{\cellcolor[HTML]{00b0f0}Automotive} & \textit{\cellcolor[HTML]{00b050}Real} & \textit{\cellcolor[HTML]{ab9ac0}Synthetic} &
\cellcolor[HTML]{a5a5a5}\textit{numFisheyeCameras} & \cellcolor[HTML]{e5b9b5}\textit{Year} & \cellcolor[HTML]{7d9ebf}\textit{Resolution} & \cellcolor[HTML]{e8715b}\textit{Tasks} \\ 
\midrule
WoodScape~\cite{yogamani2019woodscape} & \ch & \ch & \xm & 4 & 2021 & 1280$\times$966 & \begin{tabular}[c]{@{}l@{}}Semantic/Instance Seg.\\ Motion Seg.\\ 2D Bounding Boxes\end{tabular} \\
\hline
SynWoodScape~\cite{sekkat2022synwoodscape} & \ch & \xm & \ch & 4 & 2022 & 1280$\times$966 & \begin{tabular}[c]{@{}l@{}}Semantic/Instance Seg.\\ Motion Seg.\\ 2D/3D Bounding Boxes\\ Depth/Flow Estimation\\ Event Camera\\ LiDAR/IMU/GNSS \end{tabular} \\
\hline
KITTI 360~\cite{liao2021kitti} & \ch & \ch & \xm & 2 & 2021 & 1400$\times$1400 & \begin{tabular}[c]{@{}l@{}}2D/3D Semantic Seg.\\ 2D/3D Instance Seg.\\ 3D Bounding Boxes\end{tabular} \\
\hline
FisheyeCityScapes~\cite{ye2020universal} & \ch & \ch & \xm & 1 & 2020 & 600$\times$600 & Semantic Seg.\\ 
\hline
Oxford Robot Car~\cite{maddern20171} & \ch & \ch & \xm & 3 & 2016 & 1024$\times$1024 & \begin{tabular}[c]{@{}l@{}}LiDAR/IMU/GNSS\end{tabular} \\
\hline
THEODORE~\cite{scheck2020learning} & \xm & \xm & \ch & 1 & 2020 & 1024$\times$1024 & \begin{tabular}[c]{@{}l@{}}Semantic/Instance Seg. \\ 2D Bounding Boxes\end{tabular} \\
\hline
OmniScape~\cite{sekkat2020omniscape} & \xm & \xm & \ch & 2 & 2020 & 1024$\times$1024 & \begin{tabular}[c]{@{}l@{}}Semantic/Instance Seg.\\ 2D/3D Bounding Boxes\\ Depth Estimation\\ LiDAR/IMU/GNSS\end{tabular} \\
\hline
PIROPO~\cite{del2021robust} & \xm & \ch & \xm & 3 & 2020 & 800 $\times$ 600 & People Indoor Localization \\
\hline
Go Stanford~\cite{hirose2018gonet} & \xm & \ch & \xm & 2 & 2018 & 128$\times$128 & Traversability Estimation \\
\hline
Mo2Cap2~\cite{xu2019mo} & \xm & \ch & \xm & 1 & 2018 & 256$\times$256 & Human Pose Estimation \\
\hline
LMS Fisheye~\cite{eichenseer2016data} & \xm & \ch & \ch & 1 & 2016 & 1088 $\times$ 1088 & Motion Estimation \\
\hline
EgoCap~\cite{rhodin2016egocap} & \xm & \ch & \xm & 2 & 2016 & 1280$\times$1024 & \begin{tabular}[c]{@{}l@{}} Human Tracking\\ Human Pose Estimation \end{tabular} \\
\hline
LSD SLAM~\cite{caruso2015large} & \xm & \ch & \xm & 1 & 2015 & 640$\times$480 & \begin{tabular}[c]{@{}l@{}}Tracking/Mapping \end{tabular} \\
\bottomrule
\end{tabular}
\vspace{-0.4cm}
\end{table*}
% -------------------------------------------------
\subsection{Mulitask Models}
Multi-task learning (MTL) is carried out by learning commonly shared representations from multi-task supervisory signals. Since the introduction of deep learning, many dense prediction tasks, \ie, tasks that generate pixel-level predictions, have witnessed significant performance increases. These tasks are typically learned one at a time, with each task requiring the training of its own neural network. Recent MTL approaches~\cite{sistu2019neurall, chennupati2019auxnet}, on the other hand, have shown promising outcomes in terms of performance, computational complexity, and memory footprint by jointly handling many tasks via a learned shared representation.\par

For fisheye cameras, Sistu~\etal~\cite{sistu2019real} proposed a joint MTL model for learning object detection and semantic segmentation. The primary objective was to achieve real-time performance on a low-power embedded system on a chip using the same encoder for both tasks. They use a simple ResNet10-like encoder shared by both decoders to build an efficient architecture. Object detection employs the YOLO v2 decoder, whereas semantic segmentation employs the FCN8 decoder. Leang \etal explored different task weighting methods for the two-task setup on fisheye cameras \cite{leang2020dynamic}. FisheyeMultiNet~\cite{maddu2019fisheyemultinet} discusses the design and implementation of an automated parking system from the perspective of camera-based deep learning algorithms. On a low-power embedded system, FisheyeMultiNet is a real-time multi-task deep learning network that recognizes all the necessary objects for parking. The setup is a four-camera system that runs at 15fps and performs three tasks: object detection, semantic segmentation, and soiling detection.\par

Finally, a holistic real-time scene understanding for the near-field perception of the environment \textit{using cameras only} was presented in OmniDet~\cite{kumar2021omnidet}. They build a near-field perception system that constitutes a \emph{Level $3$} autonomous stack as shown in Figure~\ref{fig:qualitative-results}. With this framework's help, we can jointly understand and reason about geometry, semantics, motion, localization, and soiling from a single deep learning model comprising of six tasks at 60fps on embedded systems. Motivated by Rashed~\etal~\cite{rashed2019motion} who demonstrated that the geometric tasks like depth and motion can aid semantic segmentation, synergized cross links across tasks were implemented. Camera calibration was converted to a pixel-wise tensor and fed into the model such that it can adapt to various camera intrinsics. 
Sobh \etal ~\cite{sobh2021adversarial} studied the effect of adversarial attacks in a multi-task setup using OmniDet, which is important for safety-critical applications. The tests addressed both white and black box attacks for targeted and untargeted cases and the effect of using a simple defense strategy while attacking a task and analyzing the effect on the others.\par
% -------------------------------------------------

% Challenges
\section{Public Datasets and Research Directions}
\label{sec:challenges}

\subsection{Datasets}

Building an automotive dataset is costly and time consuming~\cite{uricar2019challenges}, it is currently the main bottleneck in progressing research in fisheye perception. In Table~\ref{tab:datasets}, we summarize the published fisheye camera datasets. WoodScape is a comprehensive dataset for 360$^\circ$ sensing around ego vehicle using four fisheye cameras. It is intended to complement the current automotive datasets where only narrow FOV images are available. Of these, KITTI~\cite{geiger2012we} was the groundbreaking dataset with different types of tasks. This is the first comprehensive fisheye automotive dataset to evaluate computer vision algorithms like fisheye image segmentation, object detection, and motion segmentation in detail~\cite{ramachandran2021woodscape}. The synthetic variant of the surround-view dataset Woodscape is SynWoodScape~\cite{sekkat2022synwoodscape}. Many of its flaws are covered and extended. The authors of WoodScape were unable to collect ground truth for pixel-level optical flow and depth because all four cameras were not available at the same time to sample different frames. This means that multicamera algorithms, which are conceivable in SynWoodScape, cannot be implemented in WoodScape. It contains 80k images with annotations from the synthetic dataset.\par

KITTI $360degree$ is a suburban dataset with a broader input modality, extensive semantic instance annotations, and precise localization to aid study in the visual, computing, and robotics fields. Compared to WoodScape, KITTI $360\degree$ differs in that it provides temporally coherent semantic instance annotations, 3D laser scans, and 3D annotations for inference in perspective and omnidirectional images. FisheyeCityScapes~\cite{ye2020universal} has proposed a seven-DoF extension, which is a virtual Fisheye data augmentation method. This method uses a radial distortion model to convert a rectilinear dataset to a fisheye dataset. It synthesizes fisheye images captured by the camera in various orientations, positions, and $f$ values, greatly improving the generalized performance of fisheye semantic segmentation. Oxford RobotCar~\cite{maddern20171} is a large-scale dataset focused on autonomous vehicles' long-term autonomy. Localization and mapping are the primary tasks of this dataset, which enables study into continuous learning for autonomous vehicles and mobile robotics.\par

THEODORE~\cite{scheck2020learning} is a large non-automotive synthetic dataset for indoor scenes containing 100,000 high resolution and 16 classes of diverse fisheye images in top-view. To achieve this, they create a 3D virtual environment of the living room, various human characters, and interior textures. The authors construct annotations for semantic segmentation, instance masks, and bounding boxes for object detection in addition to recording fisheye images from virtual environments. OmniScape~\cite{sekkat2020omniscape} dataset contains two front fisheye and catadioptric stereo RGB images mounted on the motorcycle with semantic segmentation, depth sequences, and vehicle dynamics captured by velocity, angular velocity, acceleration, and orientation. It also contains over 10,000 frames and data recorded by GTA V and CARLA that could also be extended to other simulators. Sequences were recorded in two distinct rooms utilizing both omnidirectional and perspective cameras for the PIROPO~\cite{del2021robust} (People in Indoor Rooms with Perspective and Omnidirectional cameras). The sequence depicts people in various positions, such as walking, standing, and sitting. Ground truth is point-based, and both annotated and non-annotated sequences are provided (each person in the scene is represented by a point in the center of the head). Over 100,000 annotated frames are accessible in total.\par

The Go Stanford~\cite{hirose2018gonet} dataset consists of about 24 hours of video from over 25 indoor environments. The experiment focuses on estimating traversability indoors using fisheye images. The Mo2Cap2~\cite{xu2019mo} dataset is used to estimate egocentric 3D human poses in a variety of unconstrained daily activities. This dataset aims to answer the challenge of mobile 3D posture estimation in a variety of activities such as walking, cycling, cooking, sports, and office work that take place in unrestricted real-world scenarios. Sports, animation, healthcare action recognition, motion control, and performance analysis can all benefit from these 3D postures. LMS Fisheye~\cite{eichenseer2016data} dataset aims to provide researchers with video sequences for developing and testing motion estimation algorithms developed for the fisheye camera. Both the synthetic sequence generated by Blender and the actual sequence recorded by the fisheye camera are provided.\par

EgoCap~\cite{rhodin2016egocap} is a markerless, egocentric, real-time motion capture dataset for full-body skeletal pose estimation from a lightweight stereo pair fisheye camera mounted on a helmet or virtual reality headset — optical inside-in method. The LSD-SLAM~\cite{caruso2015large} dataset is derived from a new real-time monocular SLAM approach. It is completely direct (\ie, it does not use key points/features) and creates large-scale, semi-dense maps in real-time on a laptop. Researchers can use this dataset to work on tracking (direct image alignment) and mapping. (Pixel-wise Distance Filtering) directly enables a unified omnidirectional model capable of modeling a central imaging device with an FoV of more than $180\degree$.
% -------------------------------------------------
\subsection{Research directions}

\textbf{Distortion Aware CNNs:} 
CNNs naturally exploit the translation invariance in the image grid, and it is broken in fisheye images due to spatially variant distortion. Spherical CNNs~\cite{coors2018spherenet, eder2020tangent} have been proposed, which can be directly used for spherical radial distortion models. However, automotive lenses are more complex, and the spherical model is unsuitable. It would be an interesting direction to generalize Spherical CNNs to a more complex fisheye manifold surface. Kernel transformer networks~\cite{su2019kernel} efficiently transfer convolution operators from perspective to equirectangular projections of an omnidirectional image, and it is more suitable to generalize to a fisheye image.\par
% -------------------------------------------------
\textbf{Handling Temporal Variation:}
As we discussed before, the sample complexity of an object detector is increased for a fisheye camera due to larger variability in appearance due to radial distortion. This is further exacerbated for temporal tasks, which require matching features across two frames, which could have two different distortions. For example, object tracking and reidentification are significantly more challenging in the case of fisheye cameras. Tracking a pedestrian moving from left to right of a static camera would require handling large radial distorted appearance variation. Similarly, for a static pedestrian, the horizontal and vertical motion of the camera causes large variations. It is also a challenge for the point feature correspondence problem, like tracking. One solution could be to explicitly embed the radial distortion in the feature vector, which can be leveraged for matching.\par
% -------------------------------------------------
\textbf{Bird-eye's View Perception:} 
In automated driving, it is essential to lift the detections on the image to 3D. It is typically achieved by inverse perspective mapping (IPM) \cite{Muad2004}, assuming a flat ground surface. It can also be enhanced by using depth estimation or fusion with 3D sensors \cite{mohapatra2021bevdetnet}. There is a recent trend of outputting directly in 3D using the IPM implicitly in the network~\cite{roddick2020predicting, philion2020lift}. It is typically achieved by transforming the abstract encoder features using a learnable rectification layer as an alternative to performing IPM at the input level. As CNNs have more context information and a learnable transformation can be more flexible, it works better than a pixel-wise IPM \cite{philion2020lift}. In the case of pinhole cameras, IPM is a linear transform, and it is relatively easy to design the spatial transformer of encoder features. However, for fisheye cameras, IPM is a complex non-linear operator, and it remains an open problem to directly output in bird's eye view space.\par
% -------------------------------------------------
\textbf{Multi-Camera Modeling:} 
Most of the current work in surround-view cameras treats each of the four cameras independently and performs perception algorithms. It might be more optimal to model all four surround-view cameras jointly. Firstly, it will aid detection of large vehicles (\eg, transportation trucks) visible across two or three cameras (front, left, and rear). Secondly, it eliminates the re-identification of objects seen in multiple cameras (see Figure~\ref{fig:od-fov}) and post-processing of individual detections to form a unified output like the lane model. A multi-camera model would more efficiently aggregate information and produce more optimal outputs. \cite{Pless2003} developed a classical geometric approach of treating multiple cameras as single cameras. However, there is some recent work that makes use of multiple cameras as input to a single perception model~\cite{wang2022detr3d, philion2020lift}. They make use of pinhole cameras with minimal overlapping field-of-view. It is significantly more challenging to model this for surround-view cameras.\par
% -------------------------------------------------
\begin{figure}[t]
\captionsetup{font=footnotesize, belowskip=0pt}
  \centering
  \includegraphics[width=0.75\linewidth]{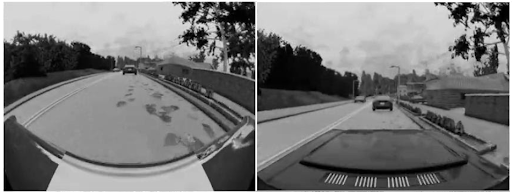}
  \caption{{\bf Illustration of near and far-field front camera images forming an asymmetric stereo pair.}}
  \label{fig:asymmetric-stereo}
\vspace{-0.4cm}
\end{figure}
% -------------------------------------------------
\textbf{Unified Modeling of Near and Far-Field Cameras:} 
A typical configuration for a next-generation automated driving system comprises full $360\degree$ coverage of near-field using four surround-view cameras and six far-field cameras (one front, one rear, two on each side)~\cite{bojarski2020nvidia}. As discussed in \S\ref{subsec:config}, they have drastically different fields-of-view and ranges. Thus, it is challenging to perform unified modeling of all the cameras extending the multi-camera modeling discussed above. Figure~\ref{fig:asymmetric-stereo} illustrates the near and far-field images of the front region. They form an asymmetric stereo pair where depth could be easily computed instead of the more challenging monocular depth, which has fundamental ambiguities. Currently, there are no public datasets containing both near and far-field cameras to enable this research.\par
% -------------------------------------------------

%-------------------------------------------------
\section{Conclusion}
\label{sec:conclusion}

Fisheye cameras are one of the most common sensors in autonomous driving systems. Despite its prevalence, there is limited understanding of it in the automotive community as it’s a specialized camera sensor, and standard algorithms do not generalize to it. This work provided a detailed account of getting started with surround-view fisheye camera development. The paper is part tutorial describing the fisheye geometry and models in detail and part survey discussing the perception algorithms developed on fisheye. We finally provide future directions to be explored.\par
% -------------------------------------------------
\section*{Acknowledgment}
The authors would like to thank Balaji Sankar Balachandaran, Jacob Roll and Louis Kerofsky from Qualcomm for providing detailed review comments.
% -------------------------------------------------
\bibliographystyle{IEEEtran}
\bibliography{bib/references}
% -------------------------------------------------
% \input{include/biographies}
% -------------------------------------------------
\end{document}